\pdfoutput=1
\documentclass[11pt]{article}

\usepackage{acl}

\usepackage{times}
\usepackage{latexsym}

\usepackage[T1]{fontenc}
\usepackage[utf8]{inputenc}

\usepackage{microtype}
\usepackage{graphicx}
\usepackage{booktabs} 
\usepackage{changepage,threeparttable} % for wide tables
\usepackage{natbib}
\usepackage[utf8]{inputenc} % allow utf-8 input
\usepackage[T1]{fontenc}    % use 8-bit T1 fonts
\usepackage{url}            % simple URL typesetting
\usepackage{booktabs}       % professional-quality tables
\usepackage{amsfonts}       % blackboard math symbols
\usepackage{nicefrac}       % compact symbols for 1/2, etc.
\usepackage{microtype}      % microtypography
\usepackage{graphicx}
\usepackage{courier}
\usepackage{amsmath}
\usepackage{amssymb}
\usepackage{bbm}
\usepackage{adjustbox}
\usepackage{pifont} % For custom symbols 
\usepackage{multirow}
\usepackage{multicol}
\usepackage{changepage,threeparttable} % for wide tables
\usepackage{makecell}
\usepackage{colortbl}
\usepackage{xspace}
\usepackage{enumitem}
\usepackage{subfigure}
\usepackage{float}
\usepackage{wrapfig}
\usepackage{algorithm}
\usepackage{algorithmicx}
\newcommand*{\images}[1]{\includegraphics[width=0.25cm,height=!]{#1}}

\title{Tapilot-Crossing: Benchmarking and Evolving LLMs \\Towards Interactive Data Analysis Agents}

\author{Jinyang Li\textsuperscript{$\clubsuit$}$^{*}$, Nan Huo\textsuperscript{$\clubsuit$}$^{*}$, Yan Gao\textsuperscript{$\heartsuit$},  \textbf{Jiayi Shi}\textsuperscript{$\spadesuit$}, Yingxiu Zhao, Ge Qu\textsuperscript{$\clubsuit$},  \\ \textbf{Yurong Wu}, \textbf{Chenhao Ma}\textsuperscript{$\spadesuit$}, \textbf{Jian-Guang Lou}\textsuperscript{$\heartsuit$}, \textbf{Reynold Cheng}\textsuperscript{$\clubsuit$}\\
  \textsuperscript{$\clubsuit$}The University of Hong Kong, 
  \textsuperscript{$\heartsuit$}Microsoft Research Asia \\
  \textsuperscript{$\spadesuit$}The Chinese University of Hong Kong, Shenzhen  \\
  \texttt{\{jl0725,huonan,quge\}@connect.hku.hk}, \texttt{ckcheng@cs.hku.hk} \\ 
  \texttt{\{yan.gao,jlou\}@microsoft.com} \\}

\begin{document}
\renewcommand{\thefootnote}{\fnsymbol{footnote}}
\maketitle
\footnotetext[1]{Equal contribution.}
\begin{abstract}
Interactive Data Analysis, a collaboration between humans and LLM agents, enables real-time data exploration for informed decision-making. The challenges and costs of collecting realistic interactive logs for data analysis hinder the quantitative evaluation of Large Language Model (LLM) agents in this task. To mitigate this issue, we introduce \textsc{Tapilot-Crossing}, a new benchmark to evaluate LLM agents on interactive data analysis. \textbf{\textsc{Tapilot-Crossing}} contains 1024 interactions, covering 4 practical scenarios: \textsc{Normal}, \textsc{Action}, \textsc{Private}, and \textsc{Private Action}. Notably, \textsc{Tapilot-Crossing} is constructed by an economical multi-agent environment, \textbf{\textsc{Decision Company}}, with few human efforts. We evaluate popular and advanced LLM agents in \textsc{Tapilot-Crossing}, which underscores the challenges of interactive data analysis. Furthermore, we propose \textbf{A}daptive \textbf{I}nteraction \textbf{R}eflection (\textbf{AIR}), a self-generated reflection strategy that guides LLM agents to \textbf{\textit{learn from successful history}}.
Experiments demonstrate that \textsc{Air} can evolve LLMs into effective interactive data analysis agents, achieving a relative performance improvement of up to 44.5\%.\footnote[2]{\url{https://tapilot-crossing.github.io/}}
\end{abstract}

\section{Introduction}

The exponential growth of the big data calls for accessible data analysis or data science techniques that cater to a range of technical backgrounds in various data-driven domains, such as healthcare, games, and entertainment \citep{data-analysis-1, data-analysis-2, data-analysis-3}. 
Recently, the fast development of Large Language Model (LLM) agents \citep{liu2023agentbench, xu2023lemur, zeng2023agenttuning, gentopia, deng2024mind2web, si2023spokenwoz} have been recevied much attention. They are capable of understanding natural language queries and generating corresponding code or analysis for data manipulation and visualization by reasoning \citep{reasoning, wei2023chainofthought, reactshunyu} and tool calls \citep{apibank, huang2024metatool, qin2023tool}. 
\begin{figure}
    \centering
    \includegraphics[width=0.47\textwidth]{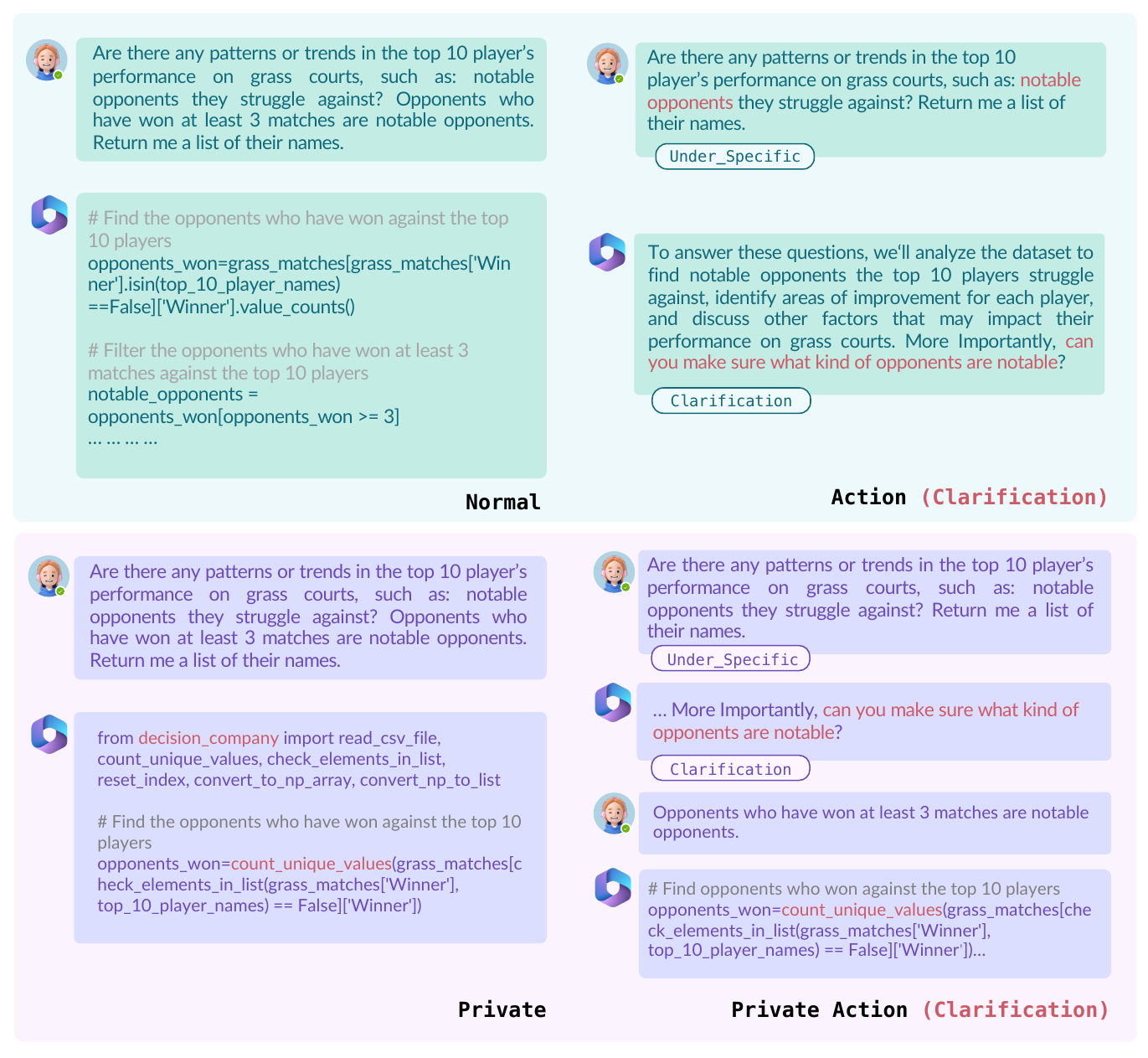}
    \caption{This is an overview of the four interaction modes in \textsc{Tapilot-Crossing}. \texttt{Notable Opponents} is ambiguous which requires clarification in multi-turn interaction.}
    \label{fig:overview}
    % \vspace{-0.5cm}
\end{figure}
SheetCopilot \citep{li2024sheetcopilot}, TableGPT \citep{zha2023tablegpt}, Data-Copilot \citep{zhang2023datacopilot} showcase the potential of tabular data analysis agents through automatic workflow given the user queries. 
However, the dynamic and uncertain nature of real-world analysis obliges effective human-agent interaction. This is because user intents can often be ambiguous \citep{devries2020ecologically, humanfbziyu, wang2024mint}, and users may need to adjust their analysis strategies based on intermediate results \citep{humanfbziyu, hcigameziyu}. 
For example, in Figure \ref{fig:overview}, the \texttt{notable opponents} could refer to a variety of interpretations, such as the opponents with the highest wins, or the most frequent opponents. 
To this end, a thorough benchmark for gauging their capability for interactive user engagement in data analysis scenarios is indispensable.

In this paper, we introduce \textsc{Tapilot-Crossing}, a new benchmark for evaluating LLM agents in interactive data analysis tasks. \textsc{Tapilot-Crossing} is designed to simulate real-world data analysis scenarios, where users interact with LLM agents to generate codes for data exploration and decision makings. It includes 1024 user-machine interactions with 1176 user intents, spanning four practical scenarios: \textbf{1) Normal}, where all questions and user requirements are explicit, requiring no actions from agents; \textbf{2) Action}, where agents must respond to diverse user feedback or instructions; \textbf{3) Private}, which examines the true semantic parsing capability of agents when encountering unseen packages during the pre-training phase \citep{private-llm}; and \textbf{4) Private Action}, a mode that combines the features of Private and Action, more closely reflecting real-world data analysis. There are two answer types: \textbf{1) Code Generation}, which can test whether the agent can correctly interpret the user's query and generate the corresponding code for data analysis, and \textbf{2) Multiple-Choice questions}, which can evaluate the agent's ability to understand the returned results being executed and provide appropriate insights for users. 

The conventional construction of datasets or benchmarks based on crowdsourcing, especially for high-quality and interactive scenarios, is time-consuming and costly due to the significant human effort and expertise required \citep{cosql, li2023llm, chase, li2024shot, zhang-etal-2023-crt}.
In this case, we design a novel multi-agent environment, \textbf{\textsc{Decision Company}}, to construct \textsc{Tapilot-Crossing}. \textsc{Decision Company} is a simulated environment where 4 GPT-4 agents communicate with each other to perform data analysis tasks. 
By using this environment, two PhD students are able to construct \textsc{Tapilot-Crossing} within a span of one months at a cost of less than 100 US dollars.

We evaluate the popular advanced LLM agents on \textsc{Tapilot-Crossing}. The results underscore the challenges of interactive data analysis and fuel the need for more advanced LLM agents that can handle diverse user intents and feedback.

To further evolve the LLMs towards effective interactive data analysis agents, we propose \textbf{A}daptive \textbf{I}nteraction \textbf{R}eflection (\textbf{AIR}), which guides LLM agents to \textbf{\textit{learn from successful history}} via self-generated pseudo logic reflection. 

Our experiments demonstrate that \textbf{AIR} can significantly enhance the performance of LLMs, in which GPT-4 can gain relative improvement of 44.5\% compared to its model base. offering an insight of how to actively improve the interaction between human and LLM agents in data analysis tasks.

In summary, our contributions are threefold:
\begin{itemize}
    \item We introduce \textsc{Tapilot-Crossing}, a new benchmark for evaluating LLM agents in interactive data analysis tasks. \textsc{Tapilot-Crossing} is constructed by our designed multi-agent environment within few human efforts, \textsc{Decision Company}, and covers a wide range of practical scenarios.
    \item We evaluate popular LLM agents on \textsc{Tapilot-Crossing}, highlighting the challenges of interactive data analysis and the need for more advanced LLM agents.
    \item We propose \textbf{\textsc{AIR}}, an effective and efficient reflection strategy that significantly improves the performance of LLM agents in interactive data analysis tasks.
\end{itemize}

\section{Preliminaries}
\begin{figure*}[t]
\centering
\includegraphics[width=1.0\textwidth]{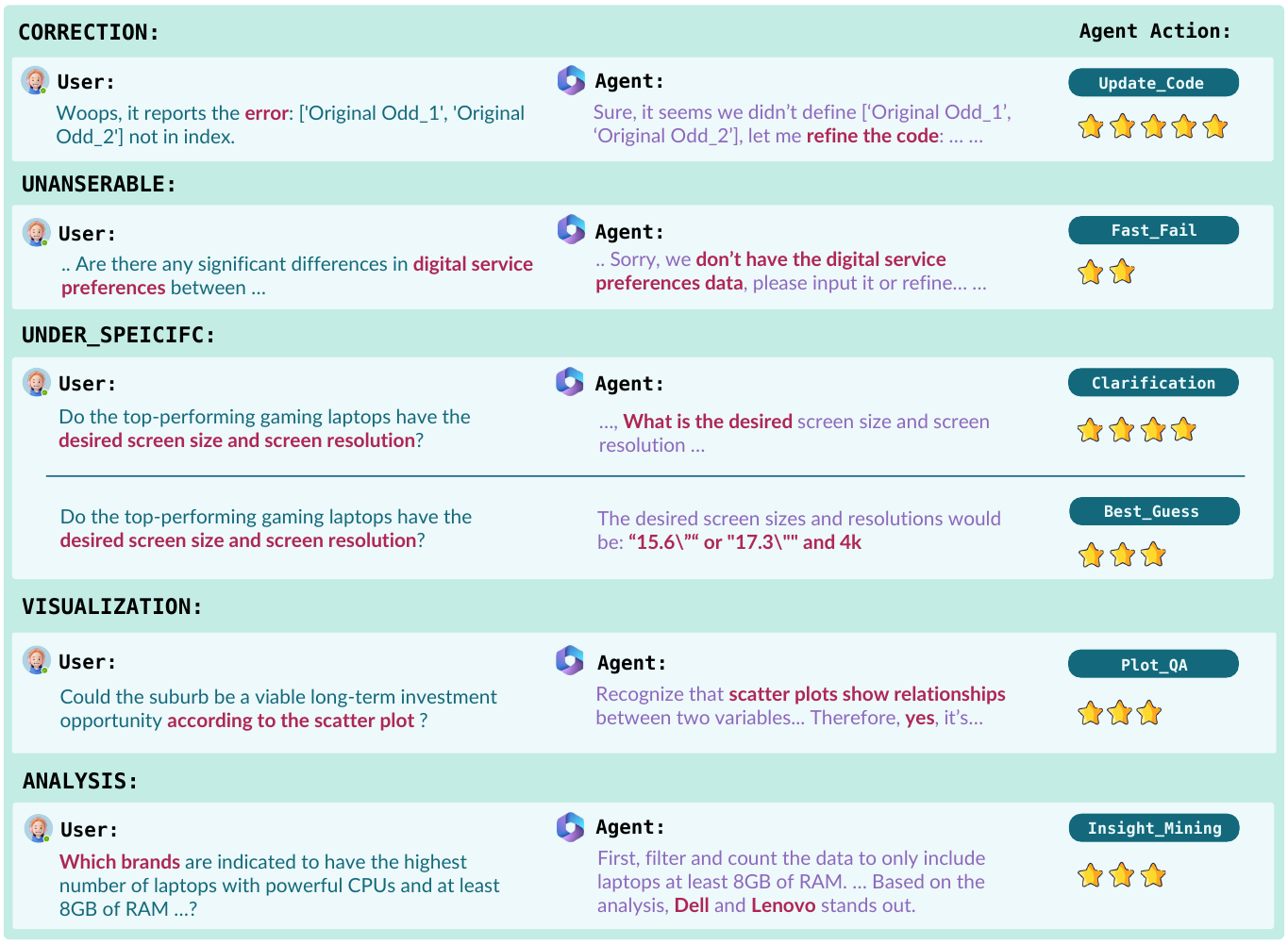}
\caption{This figure provides an overview of action types in \textsc{Tapilot-Crossing}, illustrated by examples. We emphasize the keywords specific to each category, and demonstrate the relevant sections of the associated queries, as well as the agent actions. The number of \images{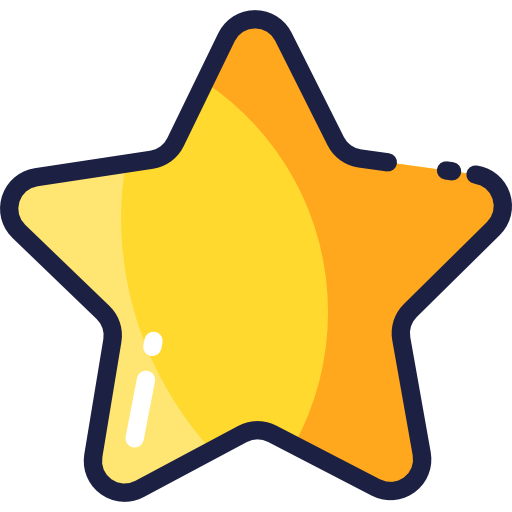} symbols represents the relative difficulty of each action.}
\label{fig:actions}
\end{figure*}
\subsection{Interactive Data analysis} The task of interactive data analysis with LLM agents involves a user and an LLM agent engaging in a sequence of user-agent turns, denoted as $[(u_1, a_1), (u_2, a_2), ..., (u_n, a_n)]$, where $n$ is the number of turns in the dialog. Each user-agent turn is a tuple $(u, a)$, where $u$ is the user's query and $a$ is the agent's response. The user's query $u$ can be a natural language instruction or a feedback to the user's or agent's previous response. The agent's response $a$ can be a code snippet for data analysis or the correct answer chosen from a list of options provided with the question. Each dialog starts with the user's initial query $u_1$ and ends with the agent's final response $a_n$. 
 
\subsection{Agent Actions in Interactions}\label{sec:action} In \textsc{Tapilot-Crossing}, we summarize 6 common agent actions in interactions to deal with the user intents and the contexts as described in Figure \ref{fig:actions}. We denote the agent action set as $A$, which is a function of the user's query $u$, the dialog history of the interaction $H$, and $T$ refers to the table contents. In this work, we mainly focus on data analysis of tabular data.

Formally, we represent the agent action set as $A = f_{\theta}(u, H, T)$, where $f_{\theta}$ refers to agent
based on LLMs with parameter $\theta$.
In \textsc{Tapilot-Crossing}, we evaluate performance of LLM agents separately in each \textsc{Action} mode of set $A$. 

\section{\textsc{Tapilot-Crossing} Dataset}\label{sec:dataset}
\begin{figure*}[t]
    \centering
    \includegraphics[width=1.0\textwidth]{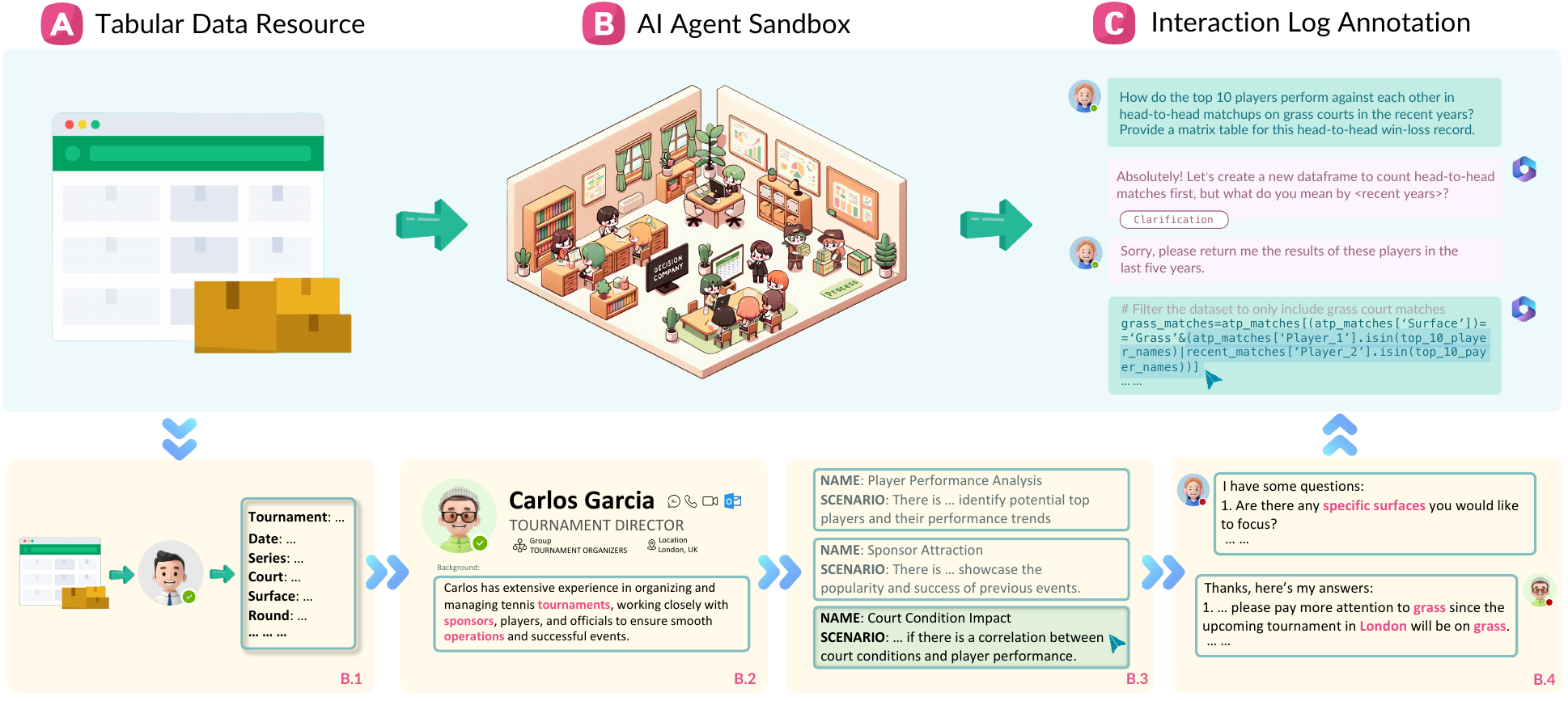}
    \caption{This describes the construction pipeline of \textsc{Tapilot-Crossing} by the AI Agent Sandbox \textsc{Decision Company}. \images{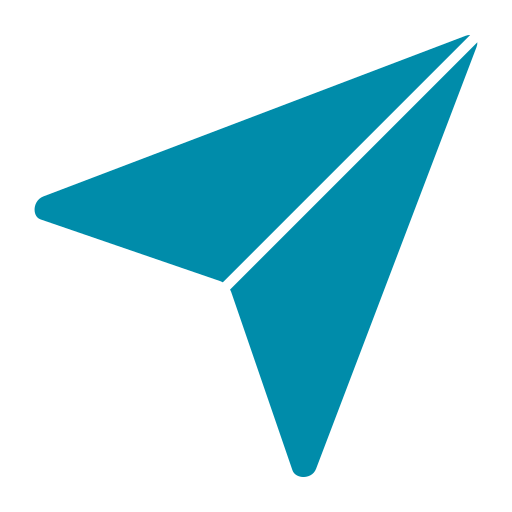} denotes human intervention during construction. For a more detailed describing, please refer to Section \ref{sec:dataset}.}
    \label{fig:decisioncompany}
\end{figure*}
\subsection{Dataset Construction.}

The construction of \textsc{Tapilot-Crossing} is mainly based on the AI Agent Sandbox, \textsc{Decision Company}, as depicted in Figure \ref{fig:decisioncompany}. \textsc{Decision Company} is a multi-agent environment where four GPT-4 agents (Administrator, Client, Data Scientist and AI Chatbot) interact with each other to perform data analysis tasks. The construction process involves the following steps: Data Acquisition \& Preprocessing, Client Persona Generation, Analysis Scenario Generation, Plan Discussion and Interaction Log Simulation.  During these stages, human intervention may be required to correct errors or eliminate harmful or biased content. Ultimately, the prototype data is adapted to private and action settings.

\paragraph{Data Acquisition \& Prepossessing.}
The first step in the construction of \textsc{Tapilot-Crossing} is the acquisition and preprocessing of data. We collect 5 open-source tables from Kaggle \footnote{\url{https://www.kaggle.com/}}, a popular data sicence platform. These datasets span diverse domains, namely \texttt{ATP Tennis}, \texttt{Credit Card}, \texttt{Fast Food}, \texttt{Laptop Price}, and \texttt{Melbourne Housing}. Then the Administrator Agent will generate column meanings and value illustrations.
\paragraph{Client Persona Generation.}
The construction of \textsc{Tapilot-Crossing} proceeds to the generation of client personas. These personas with specific tasks and topics related to the data are founded by the Administrator Agent. Each persona is defined by a \texttt{Name}, \texttt{Location}, \texttt{Job}, and \texttt{Background} with diverse range of interests and backgrounds. 
\paragraph{Simulation of Analysis Scenarios.}
Then, the Administrator Agent conducts interviews with each Client Agent to ask about their \texttt{Scenario} description, \texttt{Scenario Name}, and the \texttt{Goal} of using the dataset for the scenario. In the \textsc{Tapilot-Crossing} dataset, human annotators will interrupt here and select the most reasonable or interesting scenarios. This ensures that the scenarios included in the \textsc{Tapilot-Crossing} dataset are meaningful, not too general across different clients. For instance, in B.3 of Figure \ref{fig:decisioncompany}, we select \texttt{Court Condition Impact} because \texttt{Player Performance Analysis} is too general and \texttt{Sponsor Attraction} requires too much additional information out of the table contents, which leads to too many unawserable questions.

\paragraph{Plan Discussion.}
In this process, the Client Agent and Data Scientist Agent collaborate to convert the client's requirements into a set of specific data analysis questions. Each question is provided by an expected result type, such as dataframes, lists, or various plot types, which helps reduce question ambiguity and ease the pressure on evaluation metrics \citep{arcade, he2023text2analysis, crtqa}. The dialogue between the agents further refines the questions with specific conditions. For example, as depicted in Figure \ref{fig:decisioncompany} B.4, the client \texttt{Garcia}'s question could be further elaborated on the basis of his following responses, making all questions more answerable. In particular, Agent \texttt{ Garcia}, fully cognizant of his persona created in B.2, adds the condition \texttt{grass}, reflecting his \texttt{London} location. This implies that the role-playing aspect of the agent can be instrumental in generating a wider range of questions that are both diverse and reasonable \citep{li2024camel, aitown}.

\paragraph{Interaction Log Annotation.}
Following the plan discussion, the interaction simulation phase begins. Here, the AI Chatbot Agent takes the lead, executing the data analysis plan agreed on during the previous stage. The Chatbot Agent interacts with the Data Scientist Agent to answer a series of questions defined in the plan by generating codes and analyzing returned results. 

\subsection{Human Calibration} \label{sec:humancalibration}
While the \textsc{Decision Company} can generate a wealth of data analysis interactions in a zero-shot prompting manner, human intervention is indispensable to ensure the quality of the data set \citep{lu2023dialgen}. Our observations indicate that only 23.5\% of the original codes produced by Chatbot Agent can be directly used as reference codes. Therefore, human experts, two PhD students, participate in each stage of the generation process to calibrate the errors and meaningless interactions. While human intervention is required, it is worth noting that modifying existing answers or codes is more efficient than creating them from scratch. We preserve all natural and meaningful interactions, both agent-to-agent and human-to-machine, throughout the action setting collection.

\subsection{Private Lib Mode Evolution}
Data analyst frequently relies on their private libraries \citep{private-llm}. These libraries, often tailored to their specific needs, allow for more efficient and customized data processing and analysis. Furthermore, generating code through user-defined packages can test the true semantic parsing abilities of agents rather than merely testing their memorization of standard syntax from libraries such as Pandas\citep{ds-1000}. It also evaluates their ability to understand and implement custom functions, which is a crucial aspect of real-world data analysis. In this work, we prompt GPT-4 to autonomously convert prototype codes with pre-trained functions like Pandas or Numpy into private codes. The details can refer to Appendix \ref{sec:pri_lib_evo}.

\section{Data Statistics \& Metrics}
\subsection{Dataset Statistics}
\begin{table}[t]  
    \centering
    \resizebox{0.8\hsize}{!}{
    \begin{tabular}{lc}  
    \toprule
    \textbf{\textsc{Statistic}}& \textbf{\textsc{Number}} \\ 
    \midrule
    \textbf{Total Interactions}    & 1024  \\
    \images{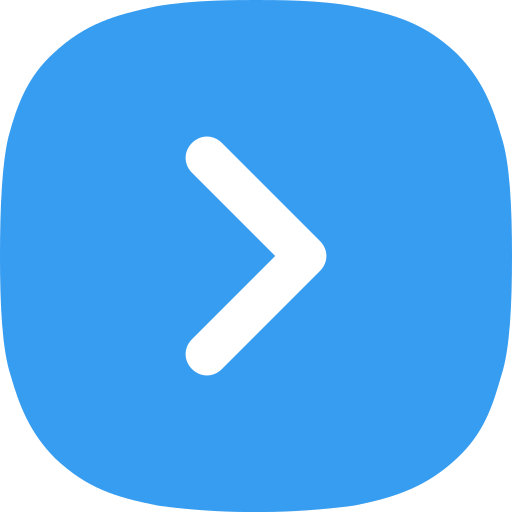} clear interactions & 284 \\
    \images{img/before.png} action interactions   & 485 \\
    \images{img/before.png} private lib. interactions & 206 \\
    \images{img/before.png} private act. interactions & 49 \\
    \images{img/before.png} \# of pirvate lib functions & 137 \\
    \midrule
    \textbf{Answer Types} &  \\
    \images{img/before.png} \# of code generation answers & 594 \\
    \images{img/before.png} \# of multi-choice answers & 430 \\
    \midrule
    \textbf{Quality \& Cost} &  \\
    \images{img/before.png} inner-agreement & 93.64 \\
    \images{img/before.png} total costs (USD) & 66.7 \\
    \bottomrule
    \end{tabular}}
    \caption{The statistics of \textsc{Tapilot-Crossing}.}
    \label{tab:statistic}
    % \vspace{-0.4cm}
\end{table}

\begin{table*}[t]
    \centering
    \renewcommand\arraystretch{1.2}
     \resizebox{\linewidth}{!}{
\begin{threeparttable}
\begin{tabular}{@{}lccccccccc@{}}
\toprule
\textbf{Dataset}   & \textbf{\# Q | \# Intents}   & \textbf{\# Toks. / Q}  & \textbf{\# Toks. / Code}  & \textbf{Code Type}   & \textbf{Analysis} & \textbf{Multi-Turn} & \textbf{Private Lib} & \textbf{Multi-modal} & \textbf{Evaluation} \\
\midrule
HumanEval \citep{chen2021evaluating} & 164 | 164    & \phantom{00}60.9 & \phantom{0}24.4 & \includegraphics[width=0.4cm]{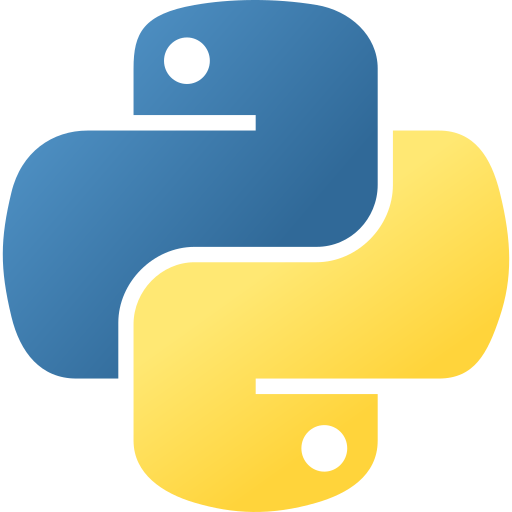}    & \includegraphics[width=0.4cm]{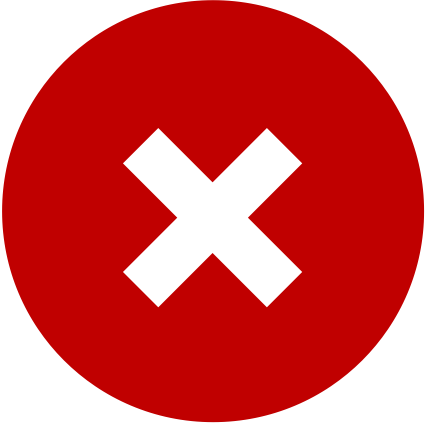} & \includegraphics[width=0.4cm]{img/no.png} & \includegraphics[width=0.4cm]{img/no.png} & \includegraphics[width=0.4cm]{img/no.png} & Test Cases \\
MBPP \citep{austin2021program}      & 974 | 974     & \phantom{00}14.5  & \phantom{0}24.2 & \includegraphics[width=0.4cm]{img/python.png}    & \includegraphics[width=0.4cm]{img/no.png} & \includegraphics[width=0.4cm]{img/no.png} & \includegraphics[width=0.4cm]{img/no.png} & \includegraphics[width=0.4cm]{img/no.png} & Test Cases \\ 
Spider \citep{spider18}       & 1034 | 1034     & \phantom{00}12.4  & \phantom{0}18.3 & \includegraphics[width=0.4cm]{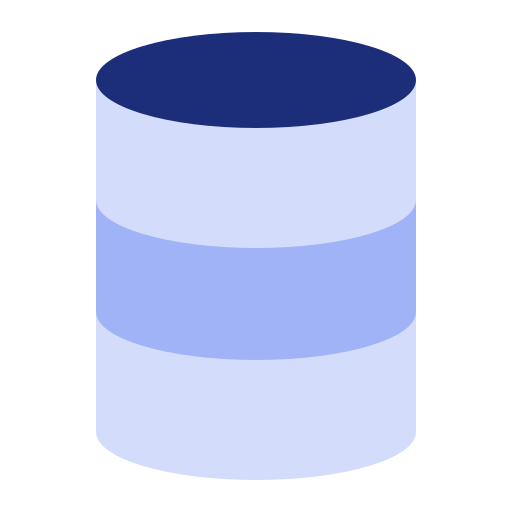}    & \includegraphics[width=0.4cm]{img/no.png} & \includegraphics[width=0.4cm]{img/no.png} & \includegraphics[width=0.4cm]{img/no.png} & \includegraphics[width=0.4cm]{img/no.png} & Acc + EM \\
BIRD  \citep{li2023llm}      & \textbf{1534} | \textbf{1534}     & \phantom{00}14.5  & \phantom{0}\underline{49.6} & \includegraphics[width=0.4cm]{img/sql.png}    & \includegraphics[width=0.4cm]{img/no.png} & \includegraphics[width=0.4cm]{img/no.png} & \includegraphics[width=0.4cm]{img/no.png} & \includegraphics[width=0.4cm]{img/no.png} & Acc + VES \\
DS-1000 \citep{ds-1000}    & 1000 | 1000  & \textbf{\phantom{0}282.4}    & \phantom{0}42.1 & \includegraphics[width=0.4cm]{img/python.png} & \includegraphics[width=0.4cm]{img/no.png} & \includegraphics[width=0.4cm]{img/no.png} & \includegraphics[width=0.4cm]{img/no.png} & \includegraphics[width=0.4cm]{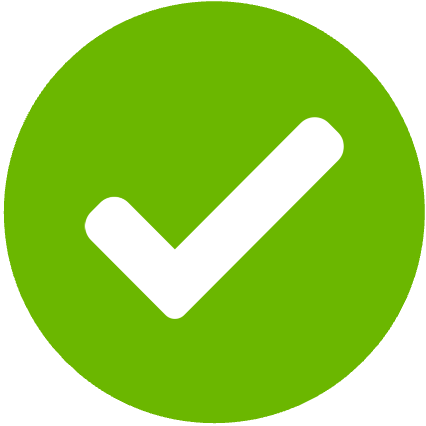} & Test Cases + SFC \\
\midrule
SparC \citep{sparc} & \underline{1203} | \underline{1203}  & \phantom{00}9.4    & \phantom{0}26.3 & \includegraphics[width=0.4cm]{img/sql.png}  & \includegraphics[width=0.4cm]{img/no.png} & \includegraphics[width=0.4cm]{img/yes.png} & \includegraphics[width=0.4cm]{img/no.png} & \includegraphics[width=0.4cm]{img/no.png} & Acc \\
CoSQL \citep{cosql} & 1008 | 1008  & \phantom{00}13.1    & \phantom{0}31.4 & \includegraphics[width=0.4cm]{img/sql.png}  & \includegraphics[width=0.4cm]{img/no.png} & \includegraphics[width=0.4cm]{img/yes.png} & \includegraphics[width=0.4cm]{img/no.png} & \includegraphics[width=0.4cm]{img/no.png} & Acc \\
ARCADE \citep{arcade}    & 1066 | 1066  & \phantom{00}19.2    & \phantom{0}48.2 & \includegraphics[width=0.4cm]{img/python.png}  & \includegraphics[width=0.4cm]{img/no.png} & \includegraphics[width=0.4cm]{img/yes.png} & \includegraphics[width=0.4cm]{img/no.png} & \includegraphics[width=0.4cm]{img/no.png} & Acc + Fuzzy \\
\midrule
\textbf{\textsc{Tapilot-Crossing}}     & 1024 | 1176  & \phantom{0}\underline{273.6}  & \textbf{110.6} & \includegraphics[width=0.4cm]{img/python.png}  & \includegraphics[width=0.4cm]{img/yes.png} & \includegraphics[width=0.4cm]{img/yes.png} & \includegraphics[width=0.4cm]{img/yes.png} & \includegraphics[width=0.4cm]{img/yes.png} & \shortstack[c]{Acc + AccR} \\
\bottomrule
\end{tabular}
\end{threeparttable}
}
\caption{
Comparison of \textsc{Tapilot-Crossing} to other data analysis datasets. The first 5 datasets are single-turn data analysis sets featuring both SQL and Python codes. The following 3 benchmarks are multi-turn or interactive data analysis datasets. \textsc{Tapilot-Crossing} represents a challenging dataset in data analysis with more comprehensive settings. \includegraphics[width=0.3cm]{img/python.png} represents that the end code is Python. \includegraphics[width=0.3cm]{img/sql.png} means the target code is SQL. 
}
\label{tab:comparison}
\end{table*}
Table \ref{tab:statistic} provides key statistics for our dataset, while Table \ref{tab:comparison} offers a comparison between \textsc{Tapilot-Crossing} and other datasets related to data analysis or science. To ensure a fair comparison regarding question and code length, we utilize \texttt{tiktoken} to compute the number of tokens for each dataset. As shown in Table \ref{tab:comparison}, \textsc{Tapilot-Crossing} includes the comprehensive evaluation settings across private library, multi-turn, and multi-modal interactions. Additionally, the complexity of this dataset, reflected in the extended lengths of questions and associated code snippets, is further amplified by the inclusion of multi-intent queries. These queries, encapsulating multiple intents within a single question, require a versatile array of computational strategies for effective handling.  For example, the query, \texttt{"Please provide mean, median, mode, range, and histogram plots for age, employment status, and credit history."} demands both statistical computations and data visualization. Finally, despite \textsc{Tapilot-Crossing} comprising 1024 data analysis interactions, it only incurs a cost of 66.7 USD, making it an economical choice for dataset generation. The inner-agreement promises the high quality of the dataset.

\subsection{Evaluation Metrics}
\paragraph{Accuracy (Acc).}
Acc is a thorough metric that evaluates ability of agents to generate code that executes correctly and to accurately answer multi-choice questions. It is defined as the proportion of instances where the predicted outputs match the expected reference output, across all evaluated tasks. For a given dataset with $N$ instances, where $C_{i}$ is the expected outcome (either execution result or correct answer) and $\hat{C}_{i}$ is the predicted outputs for the $i^{th}$ instance, Acc is calculated as follows:
\begin{equation}
\small
\mathrm{Acc}=\frac{1}{N}\sum_{i=1}^N \mathbf{I}(C_{i} = \hat{C}_{i}),
\end{equation}
where $\mathbf{I}$ is an indicator function that returns $1$ if $C_{i} = \hat{C}_{i}$, and $0$ otherwise.

\paragraph{Acc with Private Lib Recall (AccR).}
Recognizing the importance of accurately leveraging specific user-defined libraries in code generation, we extend Acc to include a recall-based adjustment for instances involving private libraries. This ensures that AccR not only evaluates the direct accuracy of code execution and question answering but also evaluates the inclusion and correct usage of private library functions. AccR can be computed as follows:
\begin{equation}
\small
\mathrm{AccR}=\frac{1}{N}\sum_{i=1}^N \mathbf{I}(C_{i} = \hat{C}_{i}) \cdot \mathbf{R}(C_{i}, \hat{C}_{i}),
\end{equation}
\begin{equation}
\small
\mathbf{R}(C{i}, \hat{C}{i})= \frac{|\mathbf{F}(C{i}) \cap \mathbf{F}(\hat{C}{i})|}{|\mathbf{F}(C{i})|},
\label{accr}
\end{equation}
where $\mathbf{R}(C_{i}, \hat{C}_{i})$ quantifies the recall rate of relevant library functions in the predicted code. $\mathbf{F}(C_{i})$ and $\mathbf{F}(\hat{C}_{i})$ denote the set of private library functions in the reference codes and the set actually utilized by agents in the predicted codes, respectively. 

\section{Evolving LLMs Towards Interactive Data Analysis Agents}
In this section, we discuss our approach of equipping LLMs as data analysis agents with tools and reasoning. Then, we introduce our self-generated reflection strategy \textsc{Air} to enhancing their performance in interactive settings.
\subsection{Toolkit}
Our tool sets include an executor, a user simulator, and a chart-to-table converter. The executor provides an environment for models to observe real-time feedback on their intermediate code results \citep{xie2023openagents, wang2024mint}. The user simulator \citep{wang2024mint, humanfbziyu}, powered by GPT-4-Turbo, tests the agents' ability to generate code after clarifying details when facing under-specific questions. The chart-to-table \citep{deplot} converter mitigates the prevalent issue of LLMs' inability to comprehend plots by converting them into tables. Detailed descriptions of these tool sets are provided in the Appendix \ref{sec:toolkit}.
\subsection{Reasoning}
Reasoning is a critical process in transitioning LLMs into data analysis agents \citep{reasoning}. In \textsc{Tapilot-Crossing}, we incorporate two primary reasoning methods for code generation and multiple-choice answers. The first is the \textbf{Chain-of-Thought (COT)} prompting technique \citep{wei2023chainofthought}, which enhances the complex reasoning abilities of LLMs by dividing the reasoning path into multiple steps. The second method is \textbf{Reasoning \& Action (ReAct)}, which enables models to make decisions by generating reasoning traces and actions in an interleaved manner, inducing writing codes, executing and understanding results, and make decision based on analysis \citep{reactshunyu}. 
\subsection{Adapative Interaction Reflection (\textsc{Air})} \label{air}
Successful interactions are important since they encapsulate the logic necessary to meet user requirements and ensure correct steps of analysis or code generation. Motivated by this, we propose an \textbf{A}daptive \textbf{I}nteraction \textbf{R}eflection (\textbf{\textsc{Air}}) approach to enable data analysis agents to learn from successful user-code histories with two steps. 
\paragraph{Pseudo Code Logic Generation.}
First, given the last previous history $(\mathbf{u}_{t-1}; \mathbf{a}_{t-1})$, when $t > 1$, we prompt the data analysis agent to reflect and generate its underlying logic $\mathbf{m}_{t-1} = f_{\theta}(\mathbf{u}_{t-1}; \mathbf{a}_{t-1})$, where $f_{\theta}$ refers to agent based on LLMs with parameter $\theta$. And $(x; y)$ represents two elements $x$ and $y$ are concatenated in the prompt. In our work, we consider pseudo code as $\mathbf{m}$ since it is an intermediate logics between natural language queries and codes.
\paragraph{Re-Org One-Shot Reasoning.}
Second, we re-organize them into a self-generated one-shot example with the order: $\mathbf{p}_{t-1}= (\mathbf{u}_{t-1}; \{\mathbf{m}_{t-1}; \mathbf{a}_{t-1}\})$, which represents the scenario where the input $\mathbf{u}_{t-1}$ is given, the agent should generate a logic $\mathbf{m}_{t-1}$ first, generate the answers $\mathbf{a}_{t-1}$. Finally, data analysis agent can learn from $\mathbf{p}_{t-1}$ to first generate logic $\mathbf{m}_{t}= f_{\theta}(\mathbf{p}_{t-1}; \mathbf{u}_{t})$ and generate answer $\mathbf{a}_{t}=f_{\theta}(\mathbf{u}_{t}; \mathbf{m}_{t})$ in the current turn $t$. 
When $t = 1$, we keep the same reasoning method of the original agent.
Figure \ref{fig:air} provides a detailed example.
\begin{table*}[t]
\centering
\resizebox{1.0\linewidth}{!}{
\begin{tabular}{ll|cccc|cc|c}
\toprule
\multirow{3}{*}{\textbf{Model}} & \multirow{3}{*}{\textbf{}} & \multicolumn{4}{c|}{\textbf{Interaction Mode}} & \multicolumn{2}{c|}{\textbf{Result Type}} & \multirow{3}{*}{\textbf{Overall}} \\
\cmidrule(lr){3-6} \cmidrule(lr){7-8}
& & \textbf{Normal} & \textbf{Action} & \textbf{Private} & \textbf{Pri-Act} & \textbf{Code} & \textbf{Choice} \\
\midrule

\multirow{3}{*}{Code-LLama-34B} & Model Base & 27.5 & 18.7 & 2.4 & 0.0 & 15.0 & 19.8 & 16.7 \\
                               & Agent      & 18.5 & 22.3 & 1.0 & 0.0 & 9.9 & 24.4 & 15.2 \\ 
                               & Inter-Agent& 28.8 & 22.9 & 2.1 & 0.2 & 15.7 & 24.8 & 19.2 \\ 
\midrule
\multirow{3}{*}{Claude-2.1} & Model Base & 20.2 & 16.8 & 1.5 & 4.5 & 11.4 & 17.9 & 13.7 \\
                            & Agent      & 23.4 & 18.0 & 3.9 & 0.0 & 13.6 & 19.1 & 15.6 \\ 
                            & Inter-Agent& 24.8 & 18.0 & 5.1 & 0.6 & 15.1 & 18.2 & 16.4 \\ 
\midrule
\multirow{3}{*}{GPT-4-Turbo} & Model Base & 27.6 & 17.5 & 5.3 & 4.3 & 17.8 & 16.1 & 17.2 \\
                             & Agent      & 29.1 & 21.7 & 10.1 & 3.6 & 20.2 & 20.6 & 20.4 \\ 
                             & Inter-Agent& 29.2 & 22.8 & \textbf{12.0} & \underline{7.9} & \underline{20.6} & 23.1 & 21.5 \\ 
\midrule
\multirow{3}{*}{GPT-4-32k} & Model Base & \underline{29.7} & 24.2 & 7.1 & 0.0 & 17.8 & 25.4 & 20.9 \\
                       & Agent      & 23.4 & \underline{39.2} & 9.1 & 5.3 & 16.6 & \underline{38.8} & \underline{25.9} \\ 
                       & Inter-Agent& \textbf{32.2} & \textbf{41.3} & \underline{10.6} & \textbf{9.8} & \textbf{21.6} & \textbf{42.1} & \textbf{30.2} \\ 
\bottomrule
\end{tabular}
}
\caption{Overall results of LLMs in base, agent, and inter-agent modes on the \textsc{Tapilot-Crossing} dataset. \textbf{Pri-Act} refers to private library +  action evaluation mode. }
\label{tab:mainresult}
\end{table*}

\section{Experiment}
We introduce experiment setup in Section \ref{exp_setup}, experimental results in Section \ref{exp_res}, and analysis in Section \ref{sec:pri-analysis} and \ref{sec:mod_dis_ana}.
\subsection{Experiment Setup}\label{exp_setup}
\paragraph{Models.} Our experiments primarily involve popular Large Language Models (LLMs) that are capable of generating code and following complex human instructions since this is a basic in data analysis. Therefore, we investigate performance of GPT-4-Turbo\footnote{\texttt{gpt-4-1106-preview}}, GPT-4-32k, Claude-2.1 and CodeLlama-34B \footnote{\texttt{codellama-34b-instruct-hf}} \citep{rozière2024codellama}.
\paragraph{Implementation details.}
Each LLM is implemented with three settings. 1) \texttt{Model-Base} refers to the LLM itself without any tool calls and reasonings. 2) \texttt{Agent} mode involves tool usage and reasoning. We employ zero-shot COT for guiding the LLM in code generation tasks since it can allow us to test the pure code generation ability of agents in data analysis. For multi-choice question answering, we utilize one-shot ReAct. 3) \texttt{Inter-Agent} mode incorporates \textsc{Air} as described in Section \ref{air} beyond the agent. Each model is provided with the last up to 5 turns of user-code histories. Further details can be found in Appendix \ref{sec:imp_detail}. For implementation of \textsc{Private} settings, we follow \citep{apibank, private-llm} to prompt agents to retrieve private libraries first then generate the code with retrieved packages.
\subsection{Experimental Results}
\begin{figure}
    \centering
    \includegraphics[width=0.35 \textwidth]{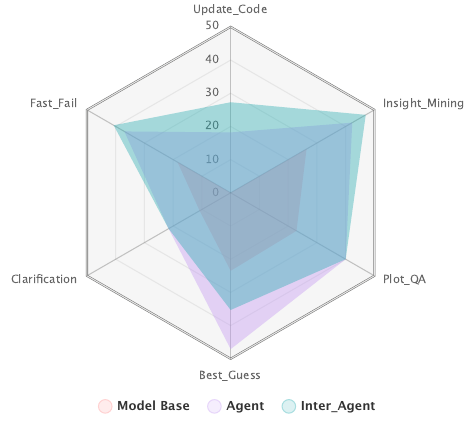}
    \caption{Visualization of the performance of GPT-4-32k across various categories in \textsc{Action} Mode. It includes a comparative analysis of the base, agent, and inter\_agent versions.}
    \label{fig:action_radar}
    \vspace{-0.8 cm}
\end{figure}
\paragraph{Overall Results.} \label{exp_res}
Table \ref{tab:mainresult} illustrates the comprehensive performance of all LLM agents and their base models on the \textsc{Tapilot-Crossing} dataset. From the results, we can deduce the following:
1) Most models with \texttt{Agent} version outperform their base version, highlighting the crucial role of tools and reasoning in enhancing the performance of Language Learning Models (LLMs) under complex tasks \citep{liu2023agentbench, xie2023openagents}.
2) All models exhibit obvious improvements in the \texttt{Inter-Agent} mode with \textsc{Air}. This indicates that the underlying logic of successful interaction histories is instrumental in guiding LLMs to become more proficient data analysis agents in interactive settings.
3) Despite GPT-4-Turbo's performance being nearly on par with GPT-4 in code generation, its overall performance still falls short of GPT-4. This indicates that beyond code writing, understanding results, and analysis are equally important. Fortunately, the comprehensive settings of \textsc{Tapilot-Crossing} can assist users in selecting models for data analysis tasks.
4) It's superising to see exceptional performance of CodeLlama in the \textsc{Normal} code generation setting. We observe that CodeLlama frequently defines functions automatically and applies these in the following code, thereby improving readability and logic. This is particularly beneficial in tasks related to data-analysis code generation. Such tasks often require the composition of API functions, which demands a profound understanding of the context and the capability to extract common patterns into reusable functions. By defining and reusing symbolic functions, CodeLlama can streamline complex contexts, making them more logical, which is an advantage for resolving complex tasks \citep{yugu2023agent}.
\paragraph{Fine-Grained Results on \textsc{Action} Modes.}
\begin{figure}
    \centering
    \includegraphics[width=0.47 \textwidth]{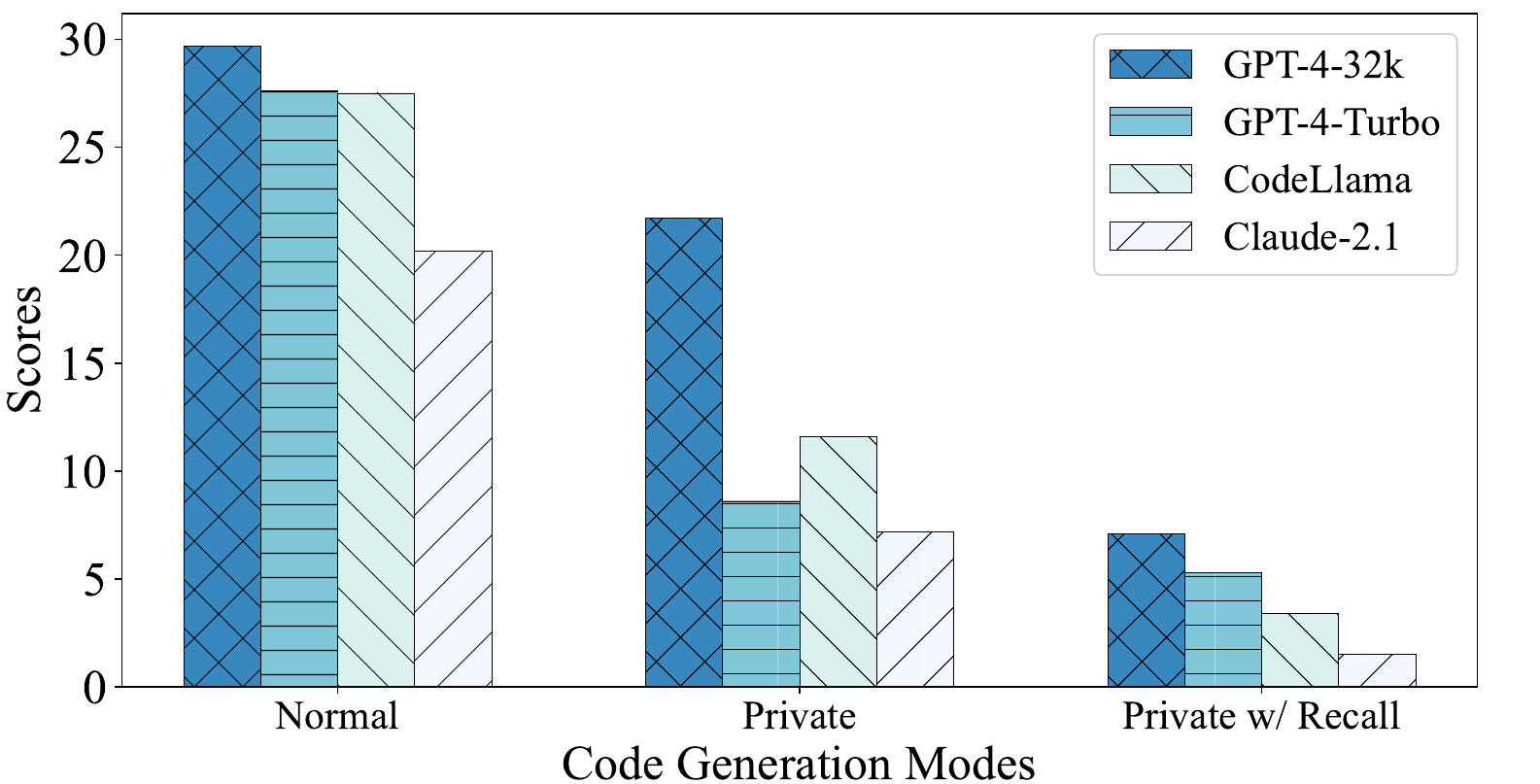}
    \caption{Visualization of the performance of LLMs on \textsc{Normal}, \textsc{Private}, \textsc{Private w/ Recall}. The Scores of first two modes are Acc. The last score is AccR.}
    \label{fig:private_lib}
    \vspace{-0.5 cm}
\end{figure}
Figure \ref{fig:action_radar} provides a comparative evaluation of three GPT-4 model variants across various \textsc{Action} modes, as detailed in Section \ref{sec:action}. The interactive data analysis agent, \texttt{Inter-Agent}, obviously outperforms in most areas, especially in managing \texttt{Fast\_Fail} queries and executing \texttt{Update\_Code} actions. However, it falls short in the \texttt{Best\_Guess} action compared to the \texttt{Agent}. We note that \textsc{Air} tends to make agents overly tractable in re-org one-shot example $\mathbf{p}_{t-1}$ and current generated logics $\mathbf{m}_{t}$. If $\mathbf{p}_{t-1}$ and $\mathbf{m}_{t}$ do not contain instructions on making assumptions, agents tend to select \texttt{None of Above}. This observation suggests that excessive reliance on historical data may hinder the inherent ability of models to conjecture based on the instant user behaviors. Therefore, striking a trade-off between user-code history exploration and real-time user interaction, especially facing under-specific questions is crucial for improving the performance of LLM agents in interactive settings.
\subsection{Private Mode Analysis} \label{sec:pri-analysis}
\paragraph{Overall Results.}
Table \ref{tab:mainresult} and Figure \ref{fig:private_lib} indicates that the \textsc{Private} setting presents a considerable obstacle, with the best performing GPT-4-Turbo \texttt{Inter-Agent} only achieving 12\%. This demonstrates that understanding and implementing user-specific functions is a critical and urgent skill for LLM agents in real-world data analysis tasks \citep{private-llm}.
\paragraph{The Critical Role of Function Relative Recall.}
Notably, CodeLlama outperforms GPT-4-Turbo in Acc within the \textsc{Private} setting. However, its performance declines significantly relative to GPT-4-Turbo upon the consideration of private library relative recall in the generated codes, as measured by the AccR metric.  This observation suggests that CodeLlama tends to reply less on user-defined private functions, aiming to reduce risk of code errors. Therefore, AccR metric can spotlight the balance required between proficient code generation and the meticulous integration of user-specified private libraries to foster safer and more satisfying code production.
\subsection{Model Shortcoming Analysis} \label{sec:mod_dis_ana}
\paragraph{Long-Context Challenges.}
The challenge of handling long-contexts is considerable in \textsc{Tapilot-Crossing}, especially for models with shorter maximum input lengths. Models such as Codellama-34B, which has a maximum input length of 16k, are particularly affected. For example, it is essential for LLMs to access all private function descriptions and codes for effective code generation with retrieved functions. The statistics shows that the average number of prompt tokens for \textsc{Private} is 15.7k, and notably, 20.4\% of their prompts surpass the 16k length.
\paragraph{Instruction Following. }
Our experiments reveal that GPT-4-32k requires minimal efforts in prompt design due to their exceptional ability to follow human instructions. To be specific, only 3.4\% of their results deviate from the provided instructions. However, other models exhibit a higher proportion of unexpected result types. For instance, extracting generated codes or answers from Claude-2.1 proves to be extremely challenging since it often embeds the answer in the middle of outputs rather than at the end as defined. We also observe that GPT-4-Turbo tends to generate longer codes in any settings. While this characteristic enhances its performance in code generation, it also results in 60.3\% of the code generated during ReAct reasoning being non-executable, thereby leading to incorrect answers. Furthermore, CodeLlama-34B-Instruct exhibits a lack of robustness when faced with longer or more complex prompts. With the addition of COT, the performance of CodeLlama significantly drops from 27.5\% with simpler instructions to 18.5\% in \textsc{Normal} code generation.

\section{Related Work}
\paragraph{Large Language Models for Data Analysis.}
The use of LLMs for data analysis has been a topic of interest in recent years. LLMs powered by In-Context Learning \citep{yang2023iterative, dai-etal-2023-gpt, dong2023survey} have been employed in various data analysis tasks, such as SQL query generation \citep{pourreza2023dinsql, gao2023dailsql, lei2023s3eval, zhang2023reactable, gu2024middleware, wang2024macsql, pourreza2024dtssql, li2024codes}, pandas or python code generation \citep{jain2023llmassisted, chen2024teaching, chen2023universal, li2024sheetcopilot, zha2023tablegpt, zhang2023datacopilot, zheng2024opencodeinterpreter}, and data visualization \citep{zhutian2023generating, huang2023lvlms}. 
However, most of these works focus on single-turn setting, where the user's query is explicit and does not require any interaction or clarification. Recently, there has been a growing interest in interactive data analysis, where the user intents may need to be clarified or refined through interactive communication \citep{devries2020ecologically, humanfbziyu, wang2024mint}. 
\paragraph{Data Analysis Benchmarks.}
The development of benchmarks for data analysis tasks has been a crucial factor in driving the progress of LLMs in data science. Existing benchmarks can be broadly categorized into single-turn and multi-turn benchmarks. Single-turn benchmarks, such as HumanEval \citep{chen2021evaluating}, MBPP \citep{austin2021program}, Spider \citep{spider18}, BIRD \citep{li2023llm}, Text2Analysis \citep{he2023text2analysis}, DABench \citep{hu2024infiagentdabench} and DS-1000 \citep{ds-1000}, focus on generating code snippets or closed-form insight summaries for data analysis given a single user query. To explore interactive nature of real-world data analysis scenarios, where the user's intent may need to be clarified or refined through interactive communication, several multi-turn benchmarks have been proposed, including CoSQL \citep{cosql}, and ARCADE \citep{arcade}. However, these benchmarks are primarily focused on code generation and do not cover other aspects of data analysis, such as data visualization and understanding based on intermediate results. Our work extends the existing literature by introducing a new benchmark, \textsc{Tapilot-Crossing}, for evaluating LLM agents in interactive data analysis tasks across wide range of data analysis settings.
\paragraph{Multi-Agent Environments for Data Generation.}
LLMs have proven to be effective in constructing multi-agent environments for automatic data generation. For instance, \citet{lu2023dialgen} and \citet{conversationgen} simulate dialogs for QA and text generation tasks. Also \citet{apibank} generates data about API calls using multi-agent environments. This is because LLM agents can simulate believable human actions when placed in an environment with dynamically updating knowledge and memory \citep{aitown}. Inspired by this, we also created \textsc{Decision Company} to generate interaction log data for data analysis with more believable behaviors. Unlike most of previous work on training dataset generation, our research pioneers the construction of the interactive benchmark with a specific focus on interactive data analysis agent evaluation. 
\section{Conclusion}
We introduce \textsc{Tapilot-Crossing}, a new benchmark for evaluating LLM agents in interactive data analysis tasks. \textsc{Tapilot-Crossing} is constructed using a cost-effective multi-agent environment, \textsc{Decision Company}, and covers a wide range of practical scenarios. We evaluate data analysis agents based on popular LLMs on \textsc{Tapilot-Crossing}, highlighting the challenges of interactive data analysis and the need for more advanced interactive data analysis agents. We also propose \textsc{AIR}, an effective reflection strategy for interactive data analysis agent evolution. Our experiments demonstrate that \textsc{AIR} can obviously enhance the performance of LLM agents. 
\newpage
\section{Limitations} \label{limitation}
\paragraph{Dataset Limitations.}
1) The \textsc{Tapilot-Crossing} dataset assumes that all human-machine interaction history is clean and correct. However, in real-world scenarios, the interaction history is often not clean, and may contain noise or require multi-turn clarifications for a single question. Therefore, future work should consider a more realistic, noisy interaction benchmark in data analysis. It is also worth noting that even with a clean history, the most capable model, GPT-4-32k, only achieves a score of 30.2 in the \texttt{Inter-Agent} mode. 2) The creation of our dataset is both cost-effective and efficient; however, the evaluation phase demands considerable efforts due to the inherent complexity and unpredictability of data analysis questions. Given it is challenging to discern subtle differences in performance among data analysis agents, especially in the context of long-form code generation and execution accuracy. This difficulty is exacerbated by the fact that the execution results of code with a single error (i.e., one-line error) and a completely incorrect code (just one-line output) are also determined as 0. Therefore, a soft-metric evaluation system should be introduced in the future as Appendix \ref{cse}. It would improve our ability to accurately gauge how close an answer is to the expected output, even when the executed output is zero, thereby providing a more fine-grained observation of code generation capabilities. 3) Finally, our work only concentrate on tabular data based analysis, while in the future, we would like to involve relational database (RDB)-based analysis with the programming language of SQLs.
\paragraph{Method Limitations.}
Our proposed reflection strategy can make LLMs more effective interactive data analysis agent, but relies heavily on the accuracy of previous interactions. This reliance becomes less reliable in instances where the historical dialogue is cluttered with errors, suggesting the need for retrieval-augmented tools or methods to identify successful past interactions. Additionally, this strategy does not enhance agent performance in initial interactions due to the absence of historical data. Finally, while effective in many \textsc{Action} settings, this focus on interaction history may limit the inferential capabilities of Large Language Models (LLMs) by prioritizing past interactions over present context. Future efforts will be directed towards refining this approach to better balance the benefits of leveraging historical interactions against the need to maintain or enhance the inferential capabilities of LLMs.
\paragraph{Model Limitations.}
In this work, we only test four popular and advanced models in comprehensive settings. Many key open-sourced models actually show limited human instruction following capability on our datasets leading to very poor performance on reasoning. Specifically, as these models are mainly trained on code, they often fail to adequately follow user instructions, as illustrated in section \ref{sec:mod_dis_ana}. These models simply return the instruction itself and continue to generate meaningless following dialogue when the resources and instructions are more complex. Therefore, how to plan more weaker LLMs to become effective interactive data analysis agents would be one of our important future work. 

\section{Ethical Statement}
The application of Large Language Models (LLMs) for automatic data generation requires  a rigorous examination of ethical implications. The primary concern is the potential for LLMs to generate contents that could be considered harmful or biased. To mitigate these risks, human annotators (two PhD students) already filter and fix all problematic cases in Section \ref{sec:humancalibration}. Also, LLMs may disseminate private or sensitive information. Therefore, we employ anonymization techniques wherein personal identifiers are systematically altered. For example, the name strings are replaced randomly, and any information of personas are switched as well. And the geographical locations of \texttt{John Smith} will be replaced with locations of \texttt{Carlos Garcia} to prevent any linkage to real-world individuals or entities. These procedures are conducted in Section \ref{sec:dataset}. Moreover, we are committed to ensuring that the outputs generated by our LLM, referred to as \textsc{Tapilot-Crossing}, are free from political or sexual biases. To this end, each output, including conclusions and generated responses, is rigorously reviewed by the authors. In a nutshell, our ethical framework is built on a foundation of transparency, accountability, and a proactive stance towards mitigating any ethical concerns associated with the use of LLMs. The measures we have implemented reflect our commitment to upholding the highest standards of ethical research practice with LLMs.

\bibliography{acl_latex}
\bibliographystyle{acl_natbib}

\newpage
\appendix
\section{License}
\subsection{\textsc{Tapilot-Crossing}}
Our Tapilot-Crossing data is available under the lisense CC BY NC 4.0.\footnote{\url{https://creativecommons.org/licenses/by-nc/4.0/}}
\subsection{\textsc{Kaggle Tabular Data}}
The tabular data that we used to create \textsc{Tapilot-Crossing} are following open-source licenses:
1) \textbf{Public Domain}: Public Domain Mark 2) \textbf{CC-BY}: Creative Commons Attribution 4.0 International.
\section{Dynamic History Combination}
\subsection{History Relational Database (H-RDB)}
\begin{figure}
    \centering
    \includegraphics[width=0.48\textwidth]{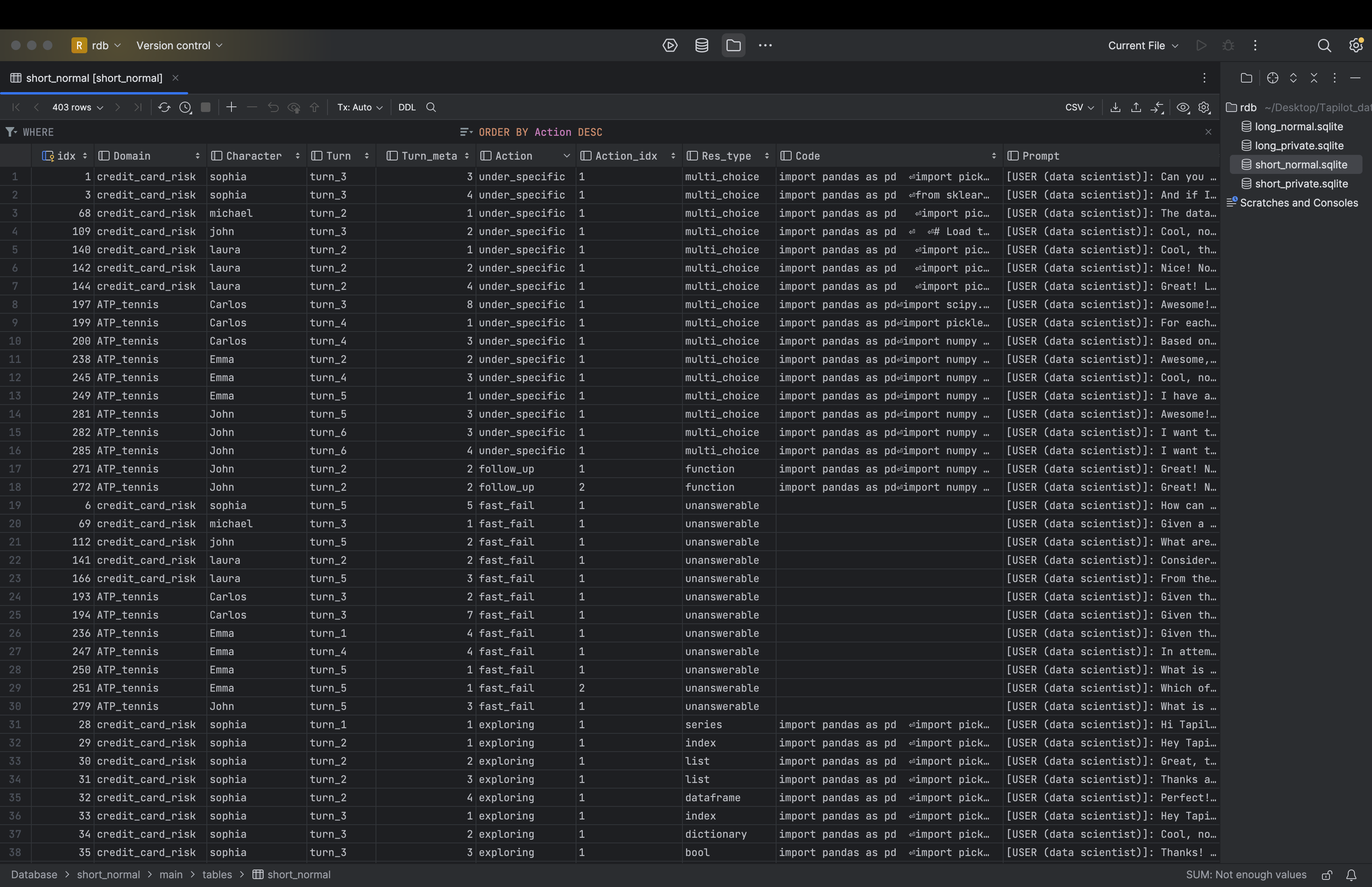}
    \caption{The screenshot of History Relational Database (H-RDB).}
    \label{fig:rdb}
    % \vspace{-0.5cm}
\end{figure}

We split the User-AI interaction into several single-turn user queries and AI answers stored in a relational database, indexed by the conversational order as shown in figure~\ref{fig:rdb}. This storage is subject to dynamic combinations for different scenarios.
\subsection{History Retrieval Queries}
When retrieving the stored history information, we use \texttt{sqlite3}\footnote{\url{https://docs.python.org/3/library/sqlite3.html}} python package. The search query is provided in sqlite3 format, for example: \texttt{SELECT \{Prompt\} FROM \{table\} WHERE 1=1 AND Domain = ? AND ...}

\subsection{Tapilot-Alpha}
As stated in Section \ref{limitation}, the current \textsc{Tapilot-Crossing} involves only clean and accurate code revisions, which we refer to as the \texttt{Alpha} version. Looking ahead, we are considering the incorporation of noisy data or the integration of user interactions in \texttt{Action} mode into the code history. This potential expansion aims to simulate more realistic development environments and challenges. Even within the constraints of a curated and error free interaction history, the experimental results show that there still are substantial opportunities for optimization and improvements.

\section{Dialog Types}
\textsc{Tapilot-Crossing} can be categorized into \textbf{Statement-} (longer) and \textbf{Colloquial-} (shorter) dialogs. The statement-dialogs are more formal, resulting in more complex user instructions and code generations, which are commonly found in computational notebooks \citep{arcade}. On the other hand, colloquial dialogs involve shorter and simpler user questions, but exhibit more colloquial and interactive characteristics. This category of dialogs is primarily constructed through the process of prompting GPT-4 to segment and reinterpret the existing statement-dialogs.

\section{\textsc{Air} Implementation}
\begin{figure*}[t]
    \centering
    \includegraphics[width=0.8\textwidth]{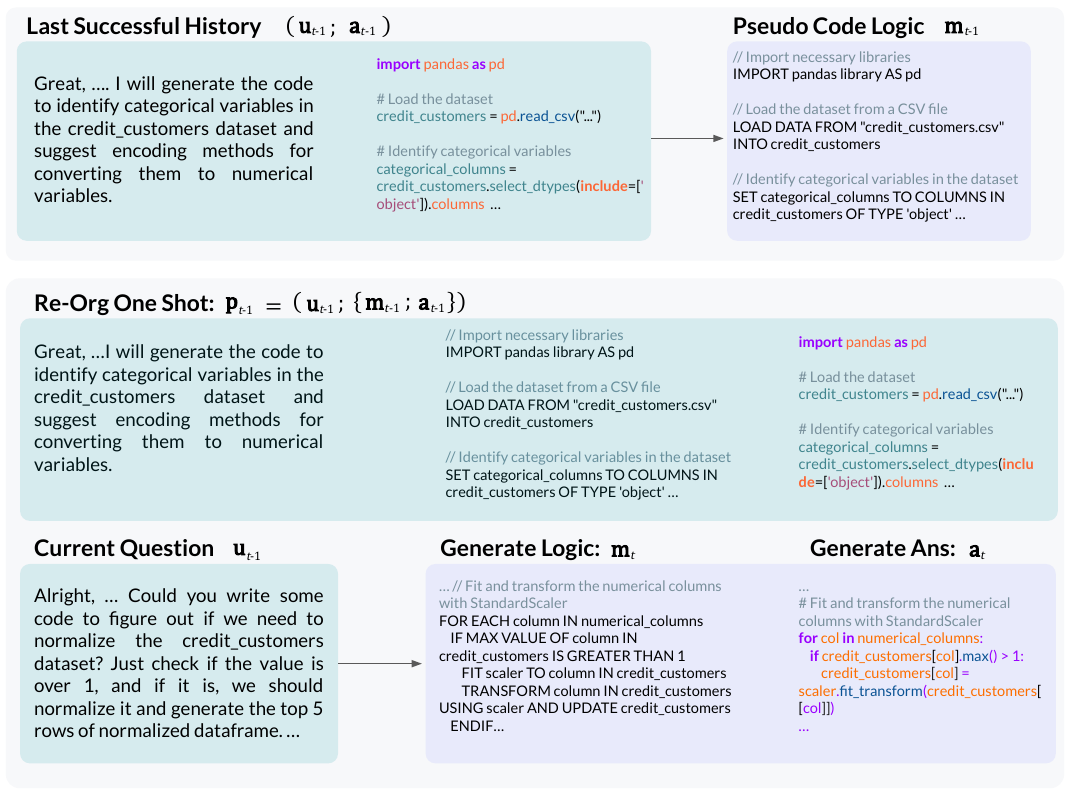}
    \caption{This is an overview of our proposed method, \textsc{Air}. The areas highlighted in purple represent results generated by the agents.}
    \label{fig:air}
\end{figure*}
Figure \ref{fig:air} presents the detailed steps of \textsc{Air}.
\section{\textsc{Agent} Implementation}
\subsection{Toolkit} \label{sec:toolkit}
\paragraph{Executor}
To get the execution results of code generated by LLMs, we adopt Python Executor \texttt{exec()} which is implemented in Python \footnote{\url{https://docs.python.org/3/library/functions.html\#exec}}, within a isolated Python environment. The output of the code execution, whether it be any return values, print statements, or error messages, is then captured by the Executor. This output is subsequently returned to the LMs, providing them with feedback on the results of their code generation to make a better next-step action or decision.

\paragraph{User Simulator}
In addressing the clarification action type, LLMs are permitted to request clarification when they feel ambigous about conditions from user queires. Therefore, we employ GPT-4 Turbo to emulate the user's question-answering behavior, considering that GPT-4 has been demonstrated to provide feedback of equivalent quality to human responses \citep{wang2024mint}.
% \begin{table}[t]  
%     \centering
%     \resizebox{0.8\hsize}{!}{
%     \begin{tabular}{lc}  
%         \toprule
%        System  & You are a helpful assistant in evaluating the similarity between two outputs generated by two different AI chatbots. Your goal is to rate the similarity between the two outputs based on a scale of 1 to 5, with 1 being highly dissimilar and 5 being highly similar.\\
%        \midrule
%        User & Rate the similarity between Output (a) and Output (b) on a scale of 1 to 5, where 1 indicates high dissimilarity, and 5 indicates high similarity. Here are some rules of the evaluation: 
                
%                     (1) Consider how closely Output (a) matches Output (b) in terms of content, context, and relevance. 
                    
%                     (2) Do not provide a rating outside the 1 to 5 scale, and avoid giving a rating of 3 (neutral) whenever possible. 
                    
%                     (3) Your judgment should be as objective as possible, without being influenced by any potential bias.
                    
%             You should answer `Score: ', followed by an integer rating between 1 to 5, where 1 indicates high dissimilarity, and 5 indicates high similarity. You should then output `Reason: ' and provide a short sentence to explain your rating. 

%             Output (a): 

%             Output (b): \\
%         \bottomrule
%     \end{tabular}}
%     \caption{The statistics of \textsc{Tapilot-Crossing}.}
%     \label{tab:statistic}
%     % \vspace{-0.4cm}
% \end{table}
\paragraph{Chart-to-Table}
We employ deplot \citep{deplot} to convert images into a table. Given the table, then LLMs can reason and answer the questions.
\subsection{Reasoning}
\paragraph{COT}
To evaluate the pure code generalization capability of data analysis, we restrict LLMs from executing code during generation. Therefore, we employ a zero-shot COT for the reasoning of the \texttt{Agent} mode.  The key prompt to implement such COT is:  \\
\texttt{... write a step-by-step outline and then write the code:}
\paragraph{ReAct}
To evaluate analytical capabilities beyond mere code generation, we employ ReAct for multiple-choice questions. Specifically, we set the \texttt{MAX STEP} for ReAct reasoning to 5, with the Executor serving as the primary tool. Data analysis agents are tasked to generate, analyze, and draw conclusions about their results. If the result contains bugs, the corresponding message is returned to the agent for rectification, although this process may consume additional reasoning steps.
We also manually provide a one-shot example to guide agents on how to react in \textsc{Tapilot-Crossing}. To prevent data leakage, we cross-reference examples across different tabular data. For instance, an example curated from \texttt{ATP\_Tennis} could be used to guide LLMs in the \texttt{Laptop Pricing} dataset.
\section{Implementation Details} \label{sec:imp_detail}
\subsection{General Implementation}
The \texttt{temperature} parameter is set to 0.0 for Claude 2.1, GPT-4, and GPT-4-Turbo.
% while a value of 0.1 is used for DeepSeek and CodeLlama. The latter setting is necessary since these open-source models would report an error if the temperature is set to 0. The \texttt{top\_p} parameter is set to 1. 

% The open-source models are implemented using \texttt{Pytorch}\footnote{https://pytorch.org/}, \texttt{Transformers}\footnote{https://huggingface.co/docs/transformers/en/installation}, and \texttt{vllm}\footnote{https://github.com/vllm-project/vllm}. To expedite the inference process, \texttt{deepspeed}\footnote{https://github.com/microsoft/DeepSpeed} is also implemented. DeepSeek and CodeLlama models are accessed via \texttt{huggingface}\footnote{https://huggingface.co/models}.
% \subsection{Private Library Retrieval for Weaker Models}
% For models with a maximum context token length of 16k, private libraries are segmented into ten sections. Agents are then tasked with selecting libraries while providing a confidence score. This approach facilitates the global selection of the most beneficial private libraries. Each model is also required to provide a reason for its confidence level and library selection. This requirement is designed to ensure that the models generate reliable function selections and confidence scores, rather than arbitrary scores based on hallucination.
\section{Action Description}
In this section, we categorize and formalize the action types in \textsc{Tapilot-Crossing}, identifying five distinct sub-categories that correspond to different types of user queries.
\subsection{Update\_Code}
The \texttt{Update\_Code} action refers to instances where the user requests corrections for bugs or refinements to the conditions of previous queries.
\subsection{Fast\_Fail}
\texttt{Fast\_Fail} is an action that alerts users when the current data contents or resources are insufficient to meet their requests, or when user queries contain factual errors.
\subsection{Clarification}
\texttt{Clarification} is a common action in response to under-specified questions, which are frequent in data-analysis queries. In this action, agents make the conditions of the question more specific and clear by seeking additional information from users.
\subsection{Best\_Guess}
While \texttt{Clarification} is an effective action to reduce the uncertainty, it can lead to issues such as user impatience due to unsteadily asking, and long dialog histories that result in attention distraction and long-context problems. Therefore, the \texttt{Best\_Guess} action can address these issues by making appropriate assumptions based on data contents, domain knowledge, and commonsense knowledge for under-specific questions. However, there is also a risk that incorrect guesses can lead to hallucinations.
\subsection{Plot\_QA}
In real data analysis settings, agents are also expected to answer user questions about insights derived from plots. The \texttt{Plot\_QA} action can assist users in better understanding the contents of plots for decision making.
\subsection{Insight\_Mining}
Beyond generating codes for users to retrieve expected results, code agents are also tasked with summarizing executed results from the environment to assist users in making informed decisions. This process, known as \texttt{Insight\_Mining}, plays an important role in data analysis since it contributes to the evolution of code agents into comprehensive data analysis agents.
\section{Evaluation Metric Details}
\subsection{DataFrame Comparison}
The function compares two dataframes (\texttt{df\_1} and \texttt{df\_2}) by checking their indices, column presence, and column data. It uses \texttt{np.allclose()} for numeric data and direct comparison for non-numeric data. If a column in \texttt{df\_1} is absent in the original dataframe, it searches for a matching column in \texttt{df\_2}. The function returns \texttt{True} if \texttt{df\_1} and \texttt{df\_2} are equivalent, otherwise \texttt{False}. Please note, the \texttt{column names} will not be computed since different LLMs may have their only preference names. For example, the \texttt{win\_ratio} generated by GPT-4 could be called \texttt{winning ratio} by Claude 2.1.
\subsection{Visualization Comparison}
We note that it's hard to compare the closed-form results for visualization-based code generation since parameters of plots may be varied significant across models. For instance, GPT-4 generated plots may be the same with CodeLlama while their title names may be different, which leads to false negatives. Therefore we utilize \texttt{PIL} package to compute similarity between plots. To be specific, the function \texttt{compare\_plots} takes two image file paths as inputs (\texttt{ai\_output} and \texttt{reference\_output}), resizes them to 800x600 pixels using the \texttt{LANCZOS} method, and saves them. The images are then read in grayscale mode to avoid the difference brought by colors. The function computes and returns the Structural Similarity Index (SSIM), a measure of image similarity, between the two images. This function can be used to compare an AI model's output with a reference output. Finally, the code generated will be considered as correct if the similarity is larger than 0.6.
\subsection{Multi-Intent Evaluation}
In this work, we evaluate the code generation performance on intent manner, which means if one user query contains multiple intents, then the total scores of this query will be the number of intents. We evaluate each intent separately and sum up the scores of all intents as the denominator when calculate the performance of each model in percentage.
\subsection{Private Function Recall}
We notice that some LLMs tend to import as many as possible private functions while not using all of them. Thus, to extract all indeed used private functions in the customized function library, we utilize \texttt{AST} package. After extracting the used private functions, we calculate the recall coefficient according to Equation \ref{accr}.
\subsection{Code Similarity Equivalance (CSE)} \label{cse}
In the context of \textsc{Tapilot-Crossing}, the complexity of code generation tasks—many of which yield a score of zero—presents huge challenges in evaluating performance through Acc or AccR only. This is particularly evident when distinguishing between codes that differ by merely a single line of error or output, both of which would result in an Acc or AccR of zero, despite their obvious differences in code generation capabilities. To overcome this limitation, we propose the introduction of Code Similarity Equivalence (CSE), a nuanced evaluation metric designed to assess the similarity between generated codes and reference codes. Given that these codes originate based on identical user instructions, a high degree of similarity is expected. Our approach leverages a hybrid combination of models to reduce the bias, incorporating CodeT5++ and OpenAI Ada (\texttt{text-embedding-ada-002}) models, which are affordable and available for most institutes. This combination has demonstrated a strong correlation with human evaluative preferences, offering a more refined and accurate measure of code generation performance.
\paragraph{Details.}
We introduce here about how to conduct more nuanced evaluation of Acc or AccR with CSE.
1) We collect 180 instances of code generation including both \textsc{Normal} and \textsc{Private}. To evaluation the quality of these codes, we enlist two additional PhD students who are proficient in data science and Python as evaluation committee.  \\
2) They evaluate code generated by several models, including GPT-4-32k, GPT-4-Turbo, Claude-2.1, CodeLlama-Instruct (ranging from 7B to 34B parameters), StarCoder, and DeepSeek-Coder-Instruct (also from 7B to 34B parameters). Each evaluator is provided with comprehensive user code histories, tabular contents, the current query, access to the \texttt{decision\_company} private library. Please note that evaluations are conducted only based on their expertise and experience, without any predefined guidelines and discussion, to avoid bias. \\
3) We ask for a relative ranking of generated codes among models over absolute scoring to avoid potential variability in scoring preferences among the evaluators. \\
4)  In cases of parts of divergent rankings, the evaluators engage in discussions regarding the specific code samples until a consensus was reached. This step ensures a more reliable and agreed-upon evaluation outcome. \\
5) The evaluation committee then examine various open-source and readily available embedding models to measure code similarity, aiming to closely match their ranking preferences. Our exploration identifies that the score system consisting of CodeT5+ \citep{wang2023codet5+} and Ada  (\texttt{text-embedding-ada-002}) most closely aligned with human evaluative preferences.
\paragraph{Introduction of a Mixed Evaluation Metric (AccSE \& AccSER).}
To accurately reflect the nuanced capabilities of code generation models, we propose a composite metric that integrates Code Similarity Evaluation (CSE) with Accuracy (Acc), termed Accuracy for Similarity Evaluation (AccSE). This metric is concisely defined as:

\begin{equation}
\small
\text{AccSE} = 
\begin{cases} 
1.0, & \text{if } C = \hat{C}, \\
0.5, & \text{if } S_1 > 0.85 \land S_2 > 0.9, \\
0.25, & \text{if } (S_1 > 0.85 \land S_2 \leq 0.9) \\
& \lor (S_1 \leq 0.85 \land S_2 > 0.9), \\
0, & \text{otherwise.}
\end{cases}
\label{accse}
\end{equation}

Where:
\begin{itemize}
    \item $C$ and $\hat{C}$ represent the reference and generated code execution outcomes, respectively.
    \item $S_1$ denotes the CSE score based on CodeT5+.
    \item $S_2$ denotes the CSE score based on Ada.
\end{itemize}

This formulation succinctly captures the evaluation criteria for AccSE, with symbols $S_1$ and $S_2$ representing the CSE scores based on CodeT5+ and Ada, respectively. The logical operators $\land$ and $\lor$ are used for "and" and "or" conditions, respectively, to further compact the notation. AccSER is computed in the similar way just times recall score for each value as Eq. \ref{accr}.

We hold this for future evaluation system of \textsc{Tapilot-Crossing} when we conduct more extensive cases with involved with more expert volunteers. 

\paragraph{Rationale Against GPT-4-Based and Multi-Agent Evaluation Methods.}
While existing research suggests that GPT-4-based soft evaluation could enhance the assessment of complex generative tasks, such approaches are deemed unsuitable for \textsc{Tapilot-Crossing} due to several critical reasons:

1) \textbf{Bias Concerns:} The prototype annotations and questions in our study originate from a GPT-4-based agent environment. Employing GPT-4 for evaluation purposes could inadvertently introduce a self-enhancement bias \citep{zheng2024judging}, compromising fairness across model evaluations.

2) \textbf{Cost Concerns:} Although multi-agent evaluation frameworks, incorporating diverse families of Large Language Models (LLMs), is to mitigate bias \citep{li2023prd}, the economical and computational overhead is obvious. Specifically, evaluations in such settings require at least twice the token consumption than that used in generation alone, rendering it impractically expensive in \textsc{Tapilot-Crossing}.

Given these considerations, our research proposes an alternative evaluation methodology that is both cost-effective and reliable for evaluating the accuracy of complex data science code generation at this time. We demonstrate that CodeT5+, a remarkably efficient code embedding model, can obviously distinguish between varying performance levels and accurately identify correct code logic. Crucially, this model offers a pragmatic balance between evaluation thoroughness and resource efficiency.

\subsection{Other Value Types}
For other result types, such as dictionry, set, list, we directly compute the exectued results and determine whether they are equal or not. 
\subsection{Case-by-Case Evaluation}
While we categorize instances according to result types and provide evaluation codes for each type, some scenarios requires a case-by-case evaluation script. For instance, in most dataframe or matrix comparisons, we employ \texttt{np.close()} and \texttt{string} match for result comparison. However, in some cases, such as using a dataframe or matrix to display a classifier's Confusion Matrix, the predicted code is deemed correct if its \texttt{f1-score} surpasses that of the referenced code, even if their \texttt{f1-scores} are not similar. For the evaluation script of \texttt{Tapilot-Crossing}, we manually review and adjust the scripts to accommodate each case.
\section{\textsc{Action} Evaluation Mode} \label{sec:action_eval_mode}
\subsection{Correction}
\paragraph{Update\_Code}
This could be evaluated within a static setting where the bug feedback is embedded into user-code history. Agents are requried to update the previous code via user feedback.
\subsection{Unawserable}
\paragraph{Fast\_Fail}
In \textsc{Decision Company}, we keep the original unanswerable questions and categorize them as multi-choice questions. This is done to evaluate if agents can identify these questions based on their analysis of table contents and commonsense knowledge. To prevent any biased setting, such as specially designed prompts that might mislead agents into determining a question as unanswerable, we sample an equal number of under-specified problems and answerable questions. We then reformulate their choices, enabling the model to decide whether a question is answerable with clarification or assumption, or to directly classify it as unanswerable.
\subsection{Under\_Specific}
\paragraph{Clarification}
To evaluate the performance of agents on clarification action, we employ a dynamic setting that incorporates a User Simulator. This simulator mimics user feedback based on the ground truth code or answer. Initially, interactive data analysis agents are expected to pose questions for clarification, simulator will answer it according to the ground truth answers. Subsequently, these agents are tasked with generating the final code, understanding both the original history and the history of clarifications. This setup provides a robust assessment of the agents' ability to interact, clarify ambiguities, and generate accurate code.
\paragraph{Best\_Guess}
We aim to evaluate the ability of interactive data analysis agents to make accurate assumptions when faced with ambiguous questions, without resorting to constant clarification, which could potentially frustrate users. We believe that an agent's best guess should not impact the final decision and this evaluation metric should be somehow flexible. For instance, in a credit card application scenario, the term \texttt{young people} could refer to individuals aged 20-40 or 25-45, making it challenging to be evaluated by fixed metrics. Therefore, we opt to use multiple-choice questions to assess the agents' assumption-making capabilities. We posit that an assumption is appropriate only if it does not influence the final decision-making process.
\subsection{Visualization}
\paragraph{Plot\_QA}
We evaluate the analysis capability of agents around plot in \textsc{Tapilot-Crossing}. The end format of answer would be multiple choices.
\subsection{Analysis}
\paragraph{Insight\_Mining}
We evaluate the analysis capability of agents generally in \textsc{Tapilot-Crossing}. We opt to use multi-choice questions to evaluate it.
\section{\textsc{Decision Company} Prompt}
\begin{figure*}[t]
    \centering
    \includegraphics[width=0.9\textwidth]{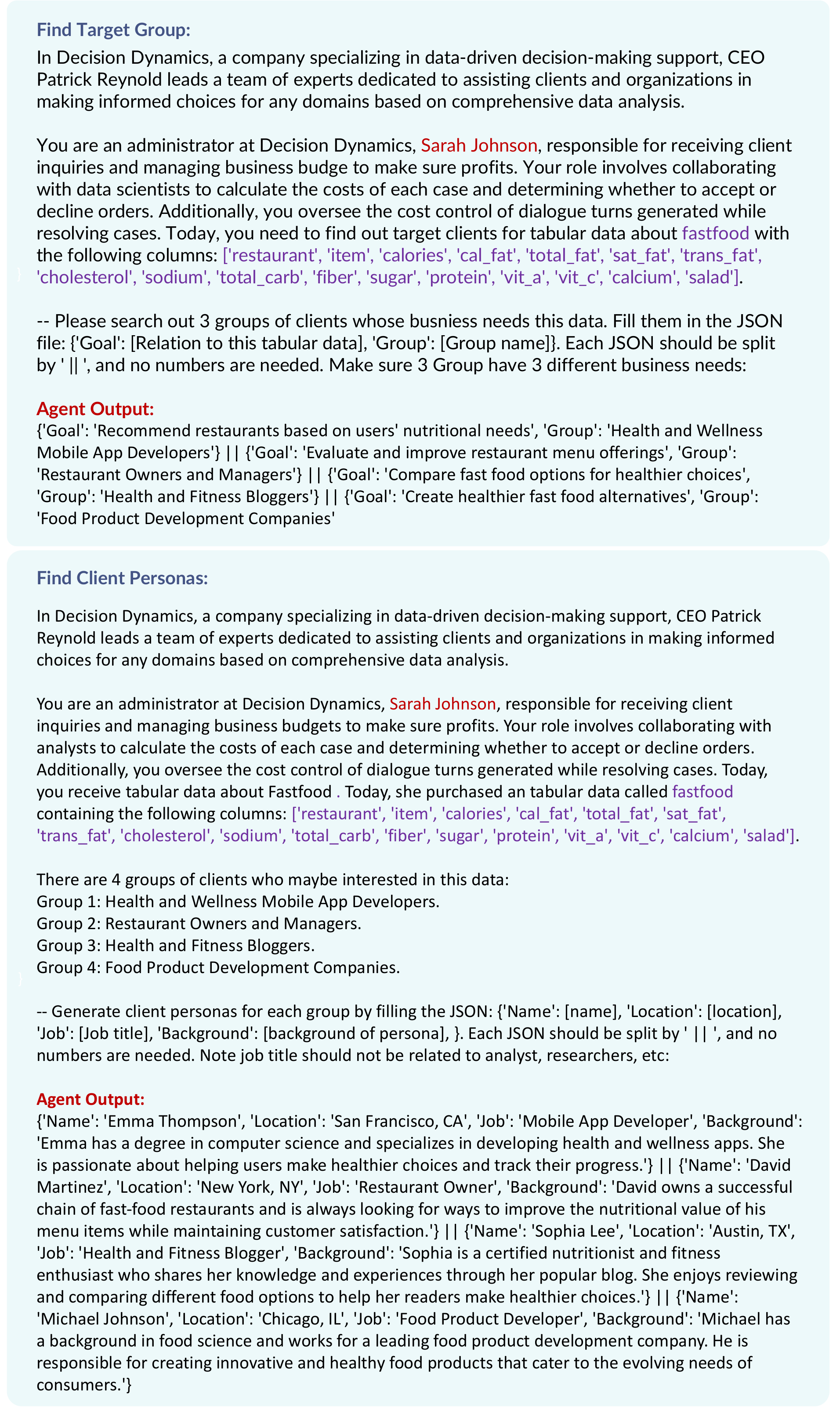}
    \caption{The prompt of Client Persona Generation}
    \label{fig:client_gen}
\end{figure*}
\begin{figure*}[h]
    \centering
    \includegraphics[width=0.9\textwidth]{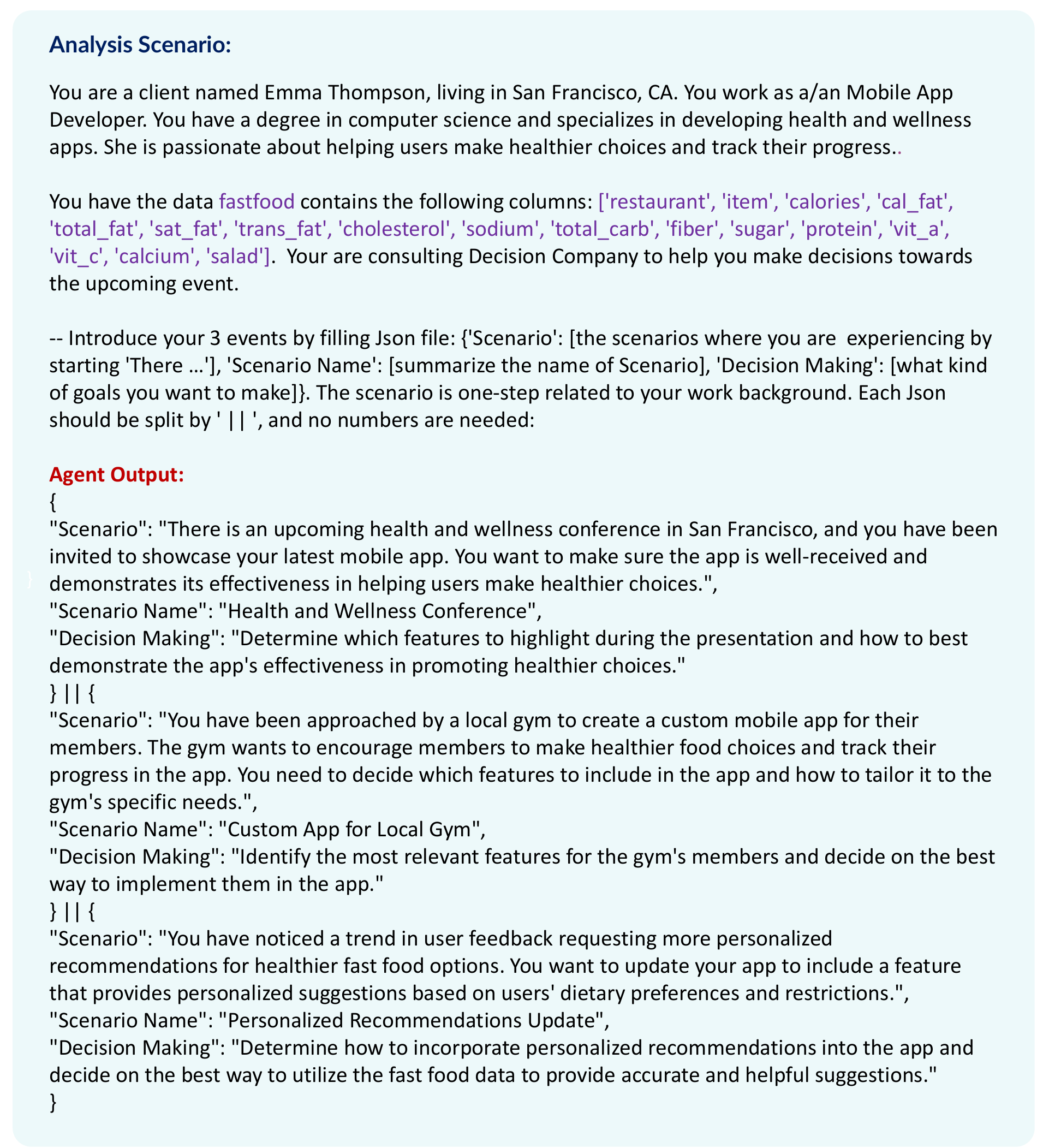}
    \caption{The prompt of analysis scenario generation}
    \label{fig:scenario_gen}
\end{figure*}
\begin{figure*}[h]
    \centering
    \includegraphics[width=1.0\textwidth]{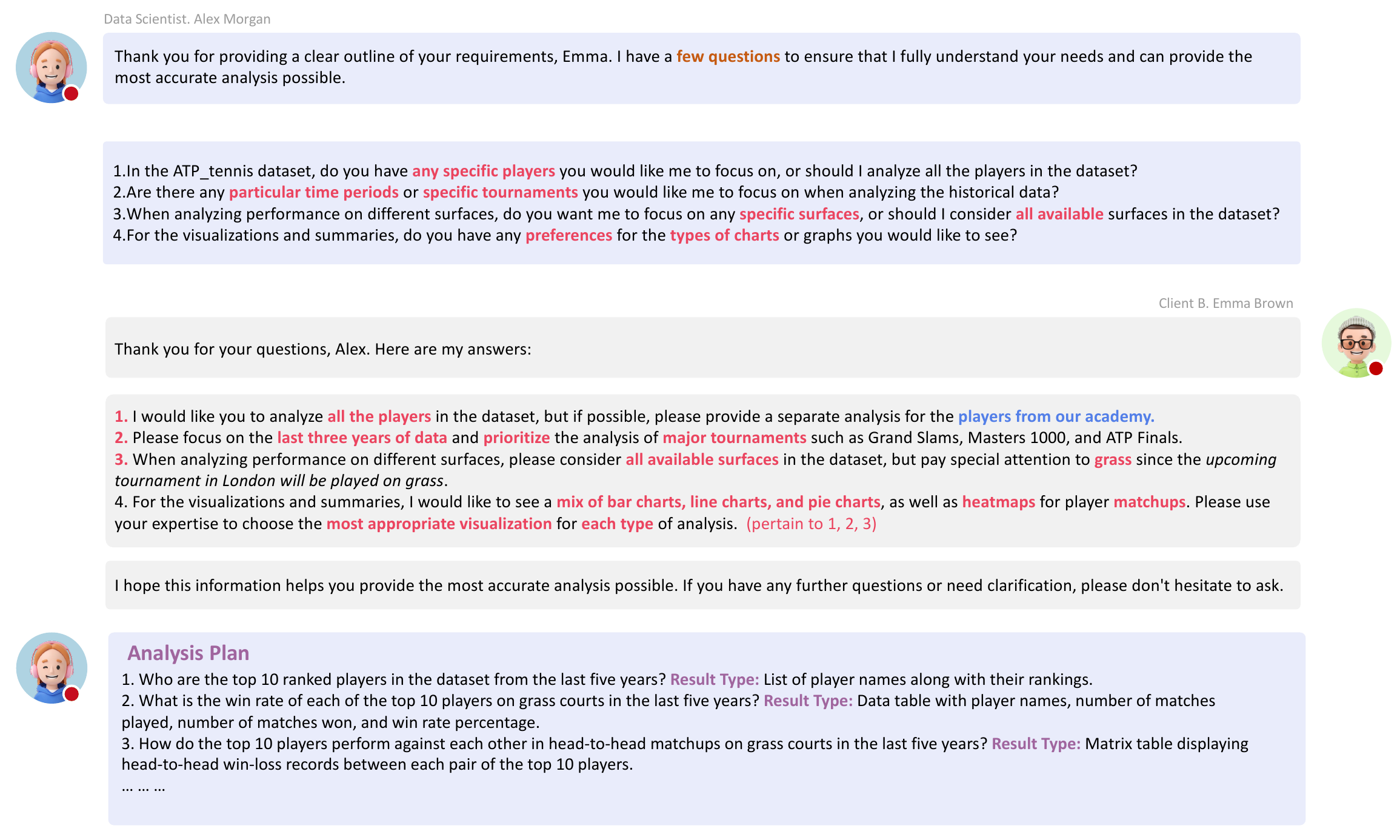}
    \caption{The example of plan discussion. The final output should be a plan of analysis invovlving questions and their or result types.}
    \label{fig:plan_discussion}
\end{figure*}
\begin{figure*}[h]
    \centering
    \includegraphics[width=0.9\textwidth]{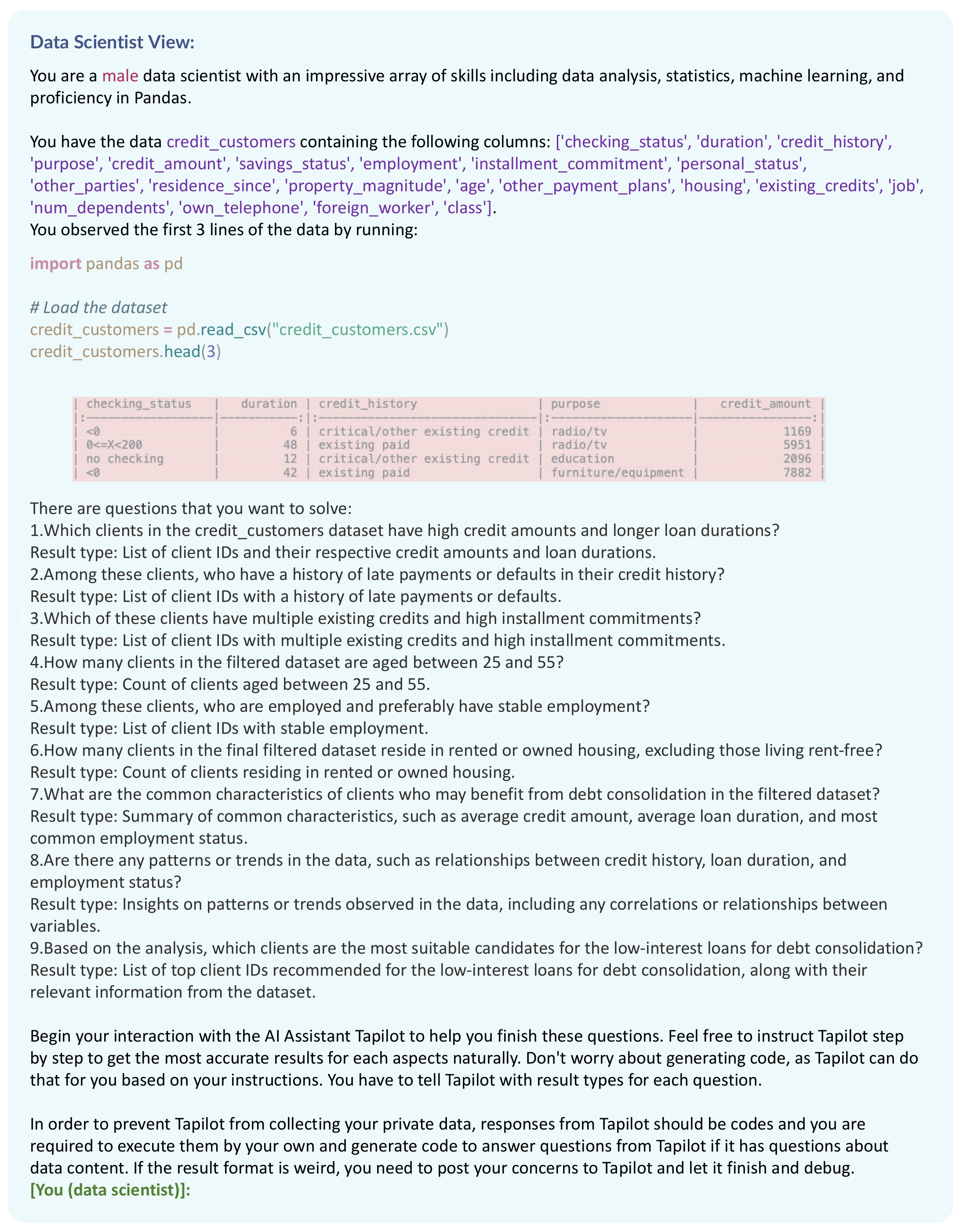}
    \caption{The prompt of Data Science Agent in interaction log generation.}
    \label{fig:ds_interaction}
\end{figure*}
\begin{figure*}[h]
    \centering
    \includegraphics[width=0.9\textwidth]{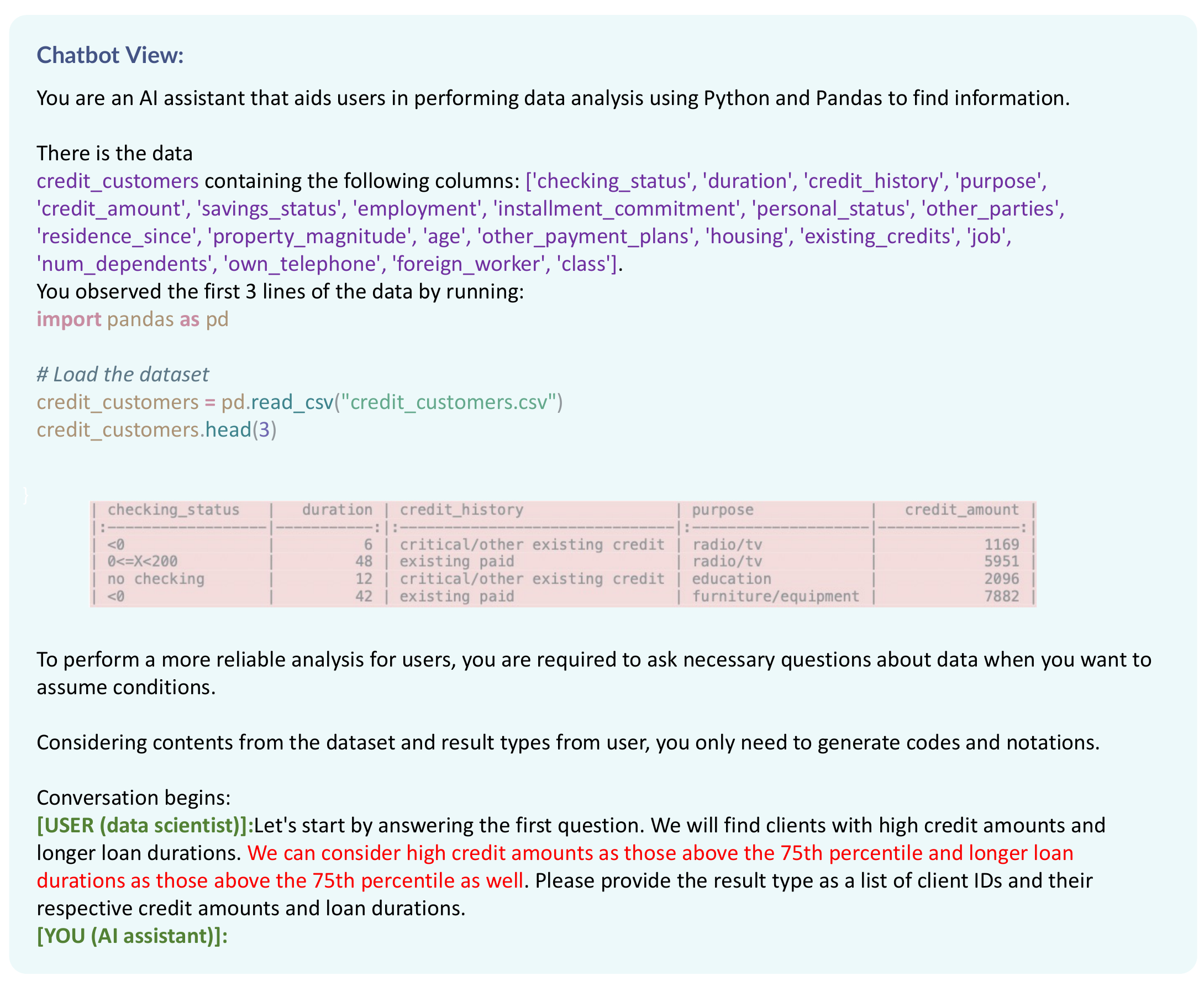}
    \caption{The prompt of Chatbot Agent in interaction log generation.}
    \label{fig:chatbot_interaction}
\end{figure*}
\begin{figure*}[h]
    \centering
    \includegraphics[width=0.9\textwidth]{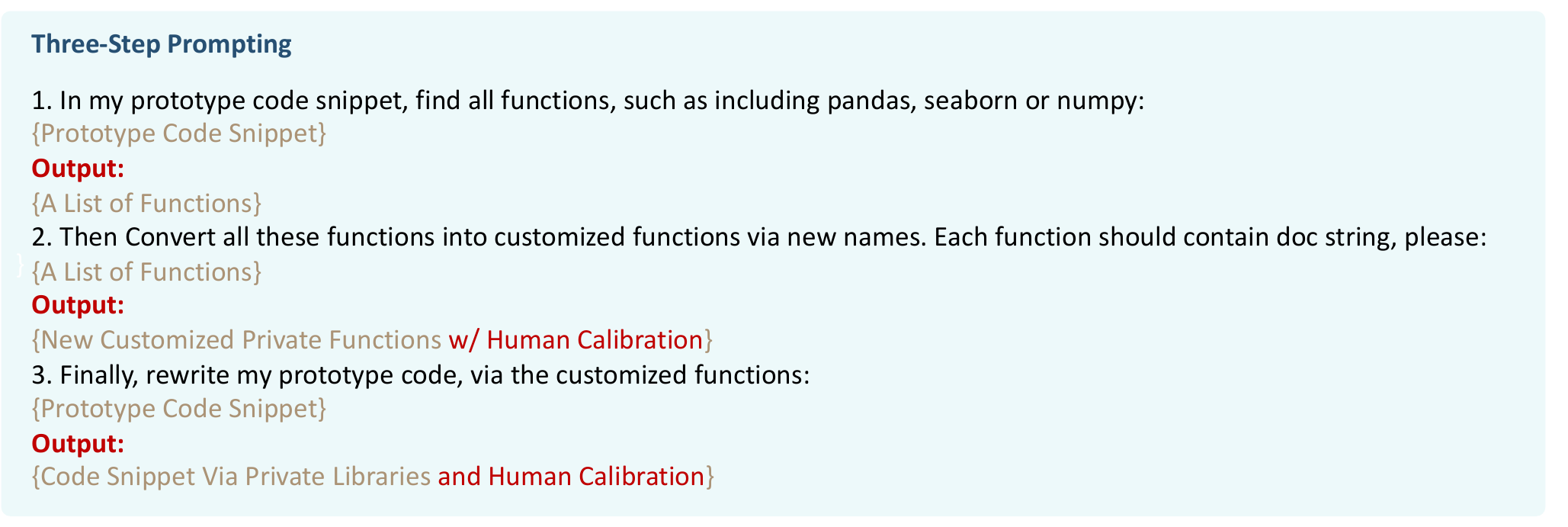}
    \caption{The prompt of Conversion from prototype code towards the code with private libraries.}
    \label{fig:private}
\end{figure*}
\subsection{Client Persona Generation}
The Figure \ref{fig:client_gen} describes how we prompt Administrator agent to generate meaningful personas.
\subsection{Simulation of Analysis Scenarios}
The Figure \ref{fig:scenario_gen} shows how we create diverse scenarios via In-Context Learning (ICL). 
\subsection{Plan Discussion}
The Figure \ref{fig:plan_discussion} presents how conversation between Data Scientist agent and Client agent can generate the series of analysis plans.
\subsection{Interaction Log Annotation}
The prompt to drive interaction log annotation is presented by Figure \ref{fig:ds_interaction} and Figure \ref{fig:chatbot_interaction} from the view of Data Scientist agent and Chatbot agent, respectively.
\subsection{Private Lib Evolution} \label{sec:pri_lib_evo}
Figure \ref{fig:private} shows the framework of how to prompt GPT-4 to generate code automatically. The human efforts are introduced to reduce the bias and fix errors.

\section{Implementation Prompt}
\subsection{\textsc{Code Generation}}
\begin{figure*}[t]
    \centering
    \includegraphics[width=0.9\textwidth]{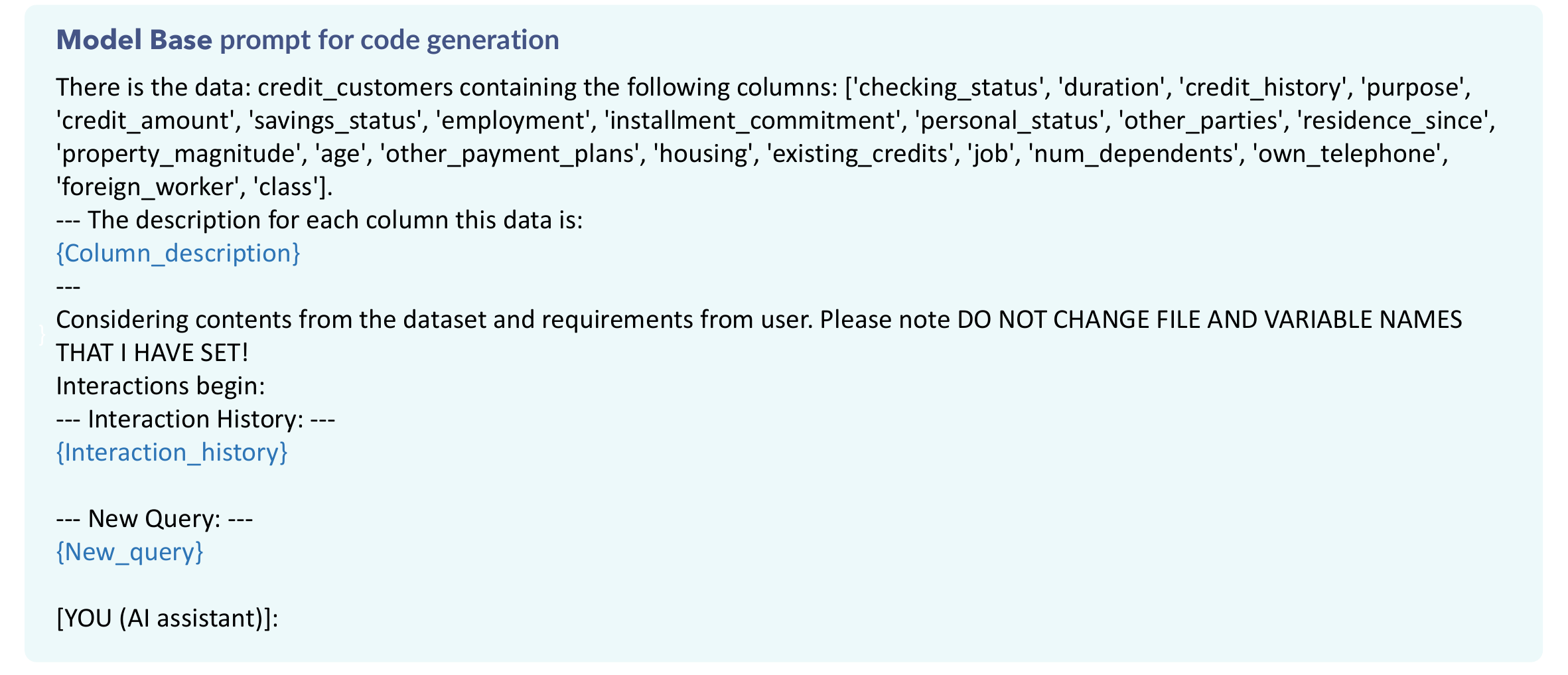}
    \caption{The prompt of LLM in \texttt{Model-Base} version in \textsc{Code Generation} mode.}
    \label{fig:prompt_base}
\end{figure*}

\begin{figure*}[t]
    \centering
    \includegraphics[width=0.9\textwidth]{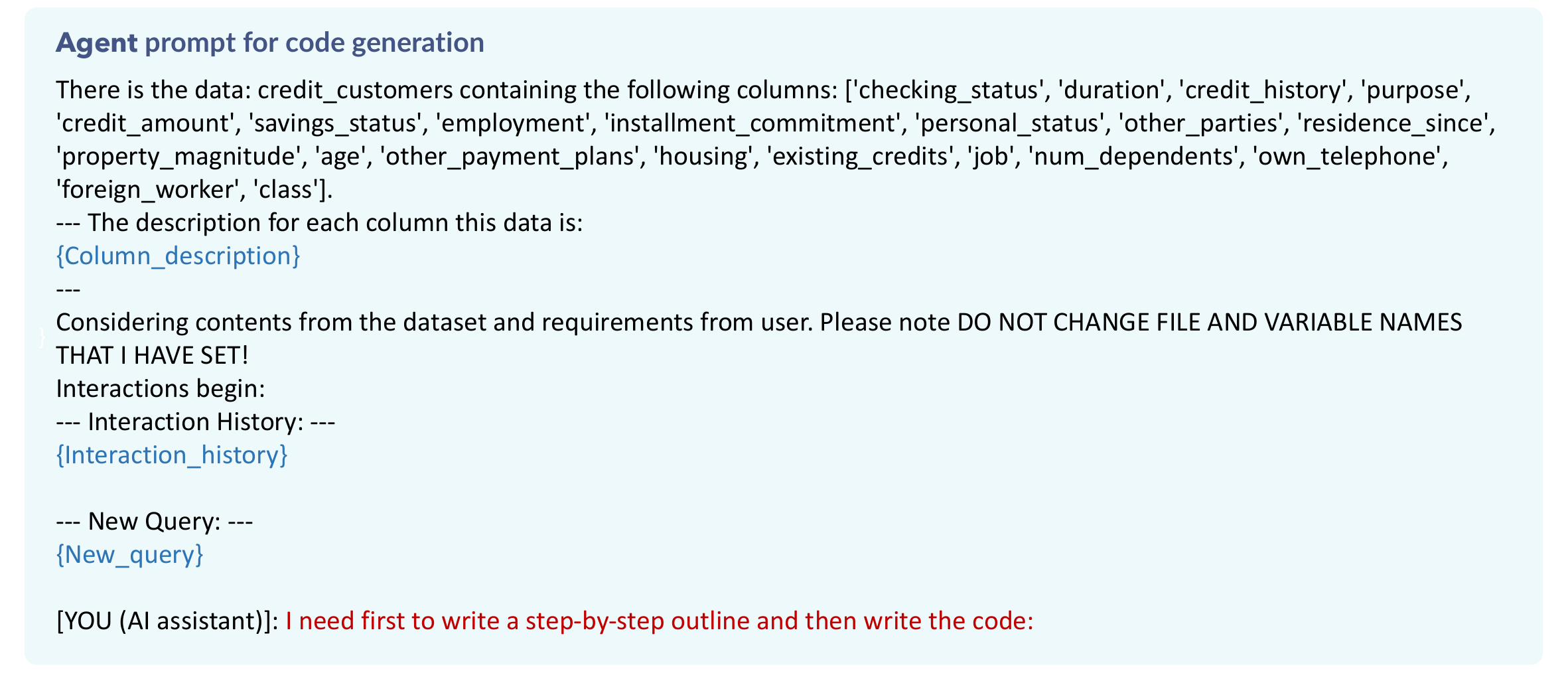}
    \caption{The prompt of LLM with data analysis agent in \textsc{Code Generation} mode. The \textbf{\textsc{COT}} prompt text is in red color.}
    \label{fig:prompt_COT}
\end{figure*}

\begin{figure*}[t]
    \centering
    \includegraphics[width=0.9\textwidth]{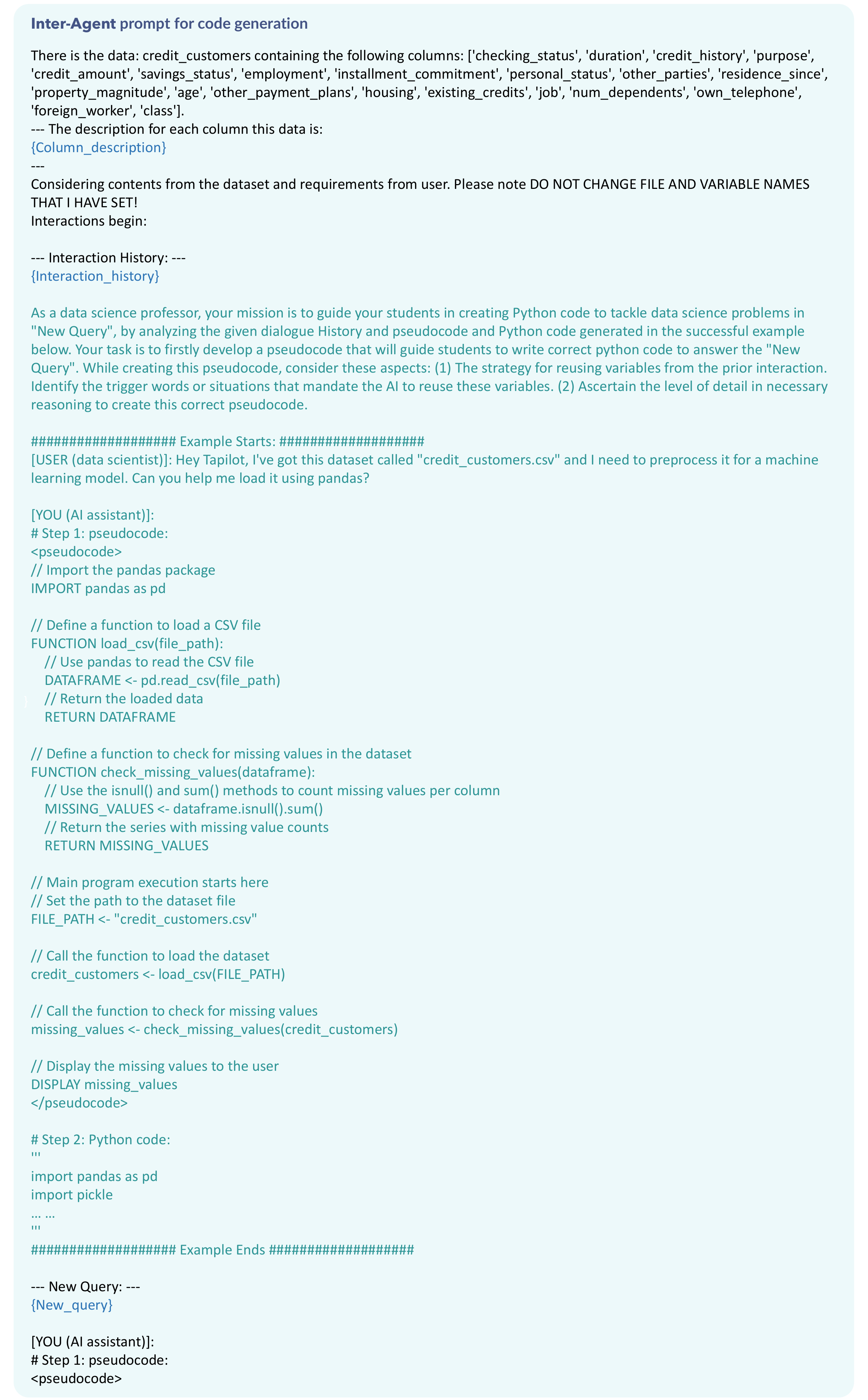}
    \caption{The prompt of LLM with interactive data analysis agent in \textsc{Code Generation} mode. The \textbf{\textsc{AIR}} prompt text are in green color, which are generated by LLM itself by learning from successful history.}
    \label{fig:prompt_AIR}
\end{figure*}

The Figure \ref{fig:prompt_base} describes how we prompt LLM model to generate code to answer user queries. And Figure \ref{fig:prompt_COT} describes how we prompt LLM in \texttt{Agent} to generate code to answer user queries following with chain-of-thought \cite{wei2023chainofthought}. Finally, Figure \ref{fig:prompt_AIR} describes how we prompt LLM in \texttt{Inter-Agent} to generate code to answer user queries with our proposed \textbf{\textsc{AIR}}.

\subsection{\textsc{Multi-choice}}
\begin{figure*}[t]
    \centering
    \includegraphics[width=0.9\textwidth]{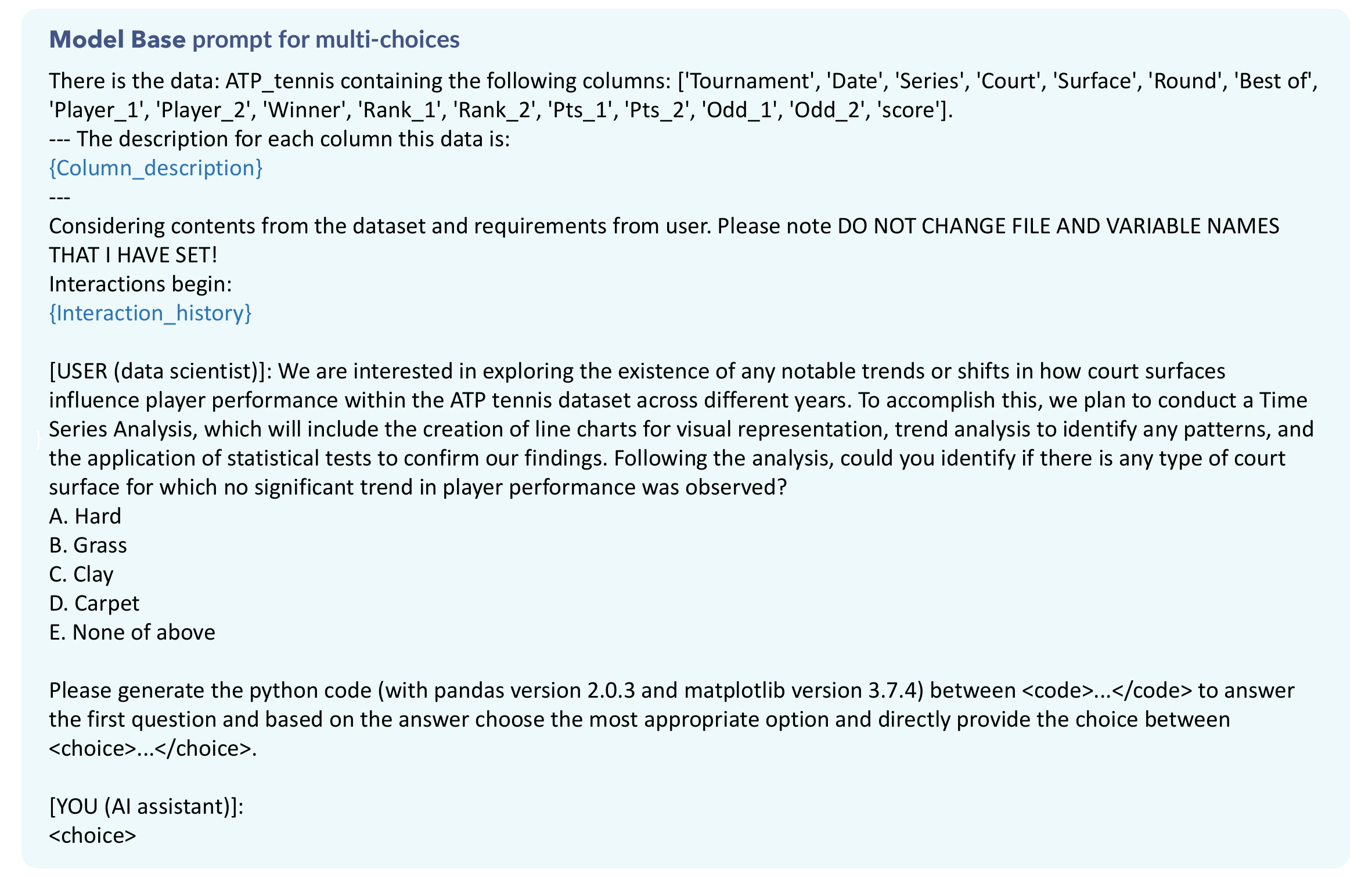}
    \caption{The prompt of LLM in \texttt{Model-Base} version in \textsc{Multi-choice} mode.}
    \label{fig:prompt_base_MC}
\end{figure*}

\begin{figure*}[t]
    \centering
    \includegraphics[width=0.9\textwidth]{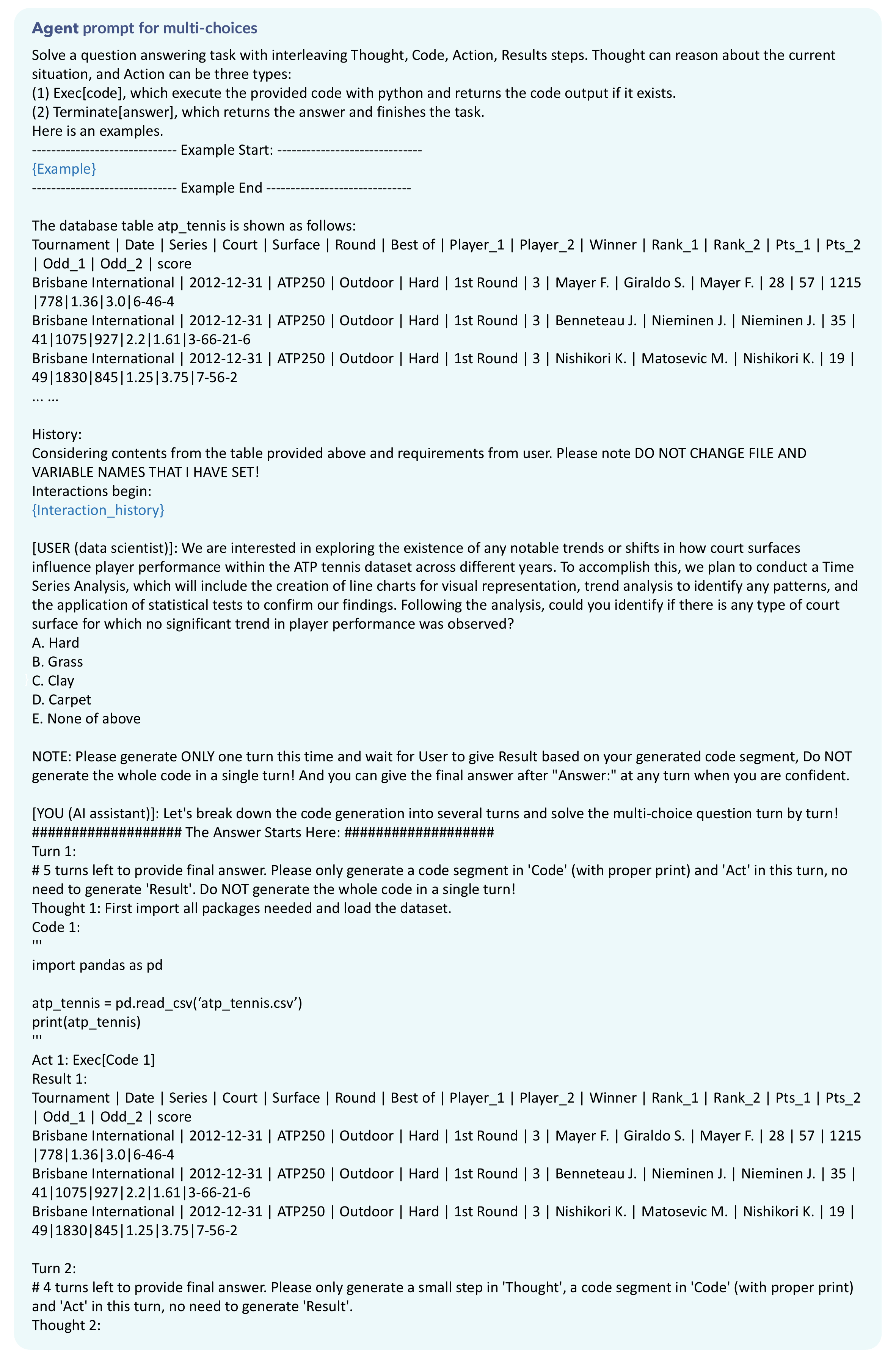}
    \caption{The prompt of LLM with data analysis agent in \textsc{Multi-choice} mode. }
    \label{fig:prompt_react}
\end{figure*}

\begin{figure*}[t]
    \centering
    \includegraphics[width=0.9\textwidth]{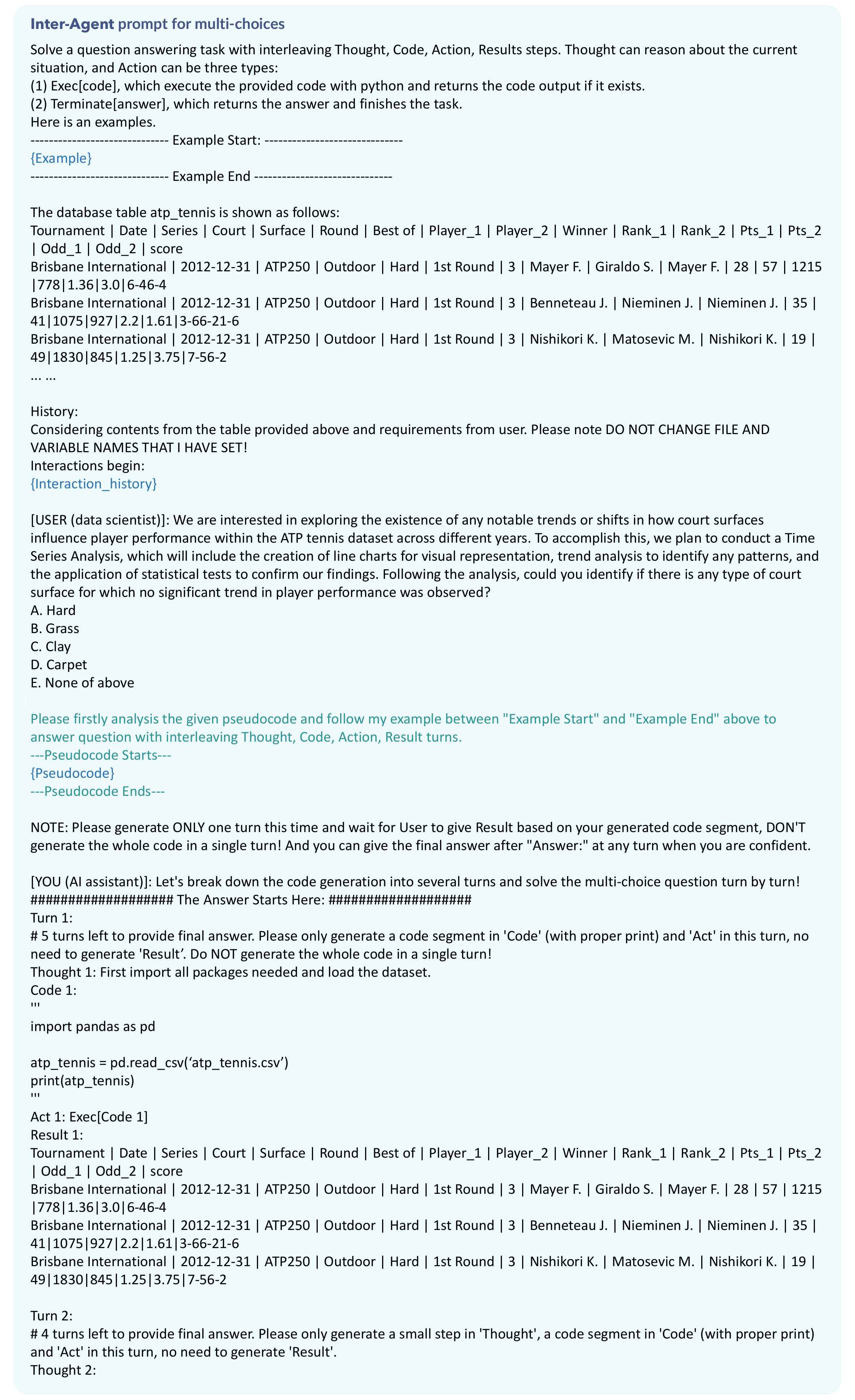}
    \caption{The prompt of LLM with interactive data analysis agent in \textsc{Multi-choice} mode. The \textbf{\textsc{AIR}} prompt text are in green color. And the pseudocode is generated by LLM itself by learning from successful history}
    \label{fig:prompt_AIR_MC}
\end{figure*}

The Figure \ref{fig:prompt_base_MC} describes how we prompt LLM to answer user queries. And Figure \ref{fig:prompt_react} describes how we prompt LLM in \texttt{Agent} to answer user queries following with ReAct \cite{reactshunyu}. Finally, Figure \ref{fig:prompt_AIR_MC} describes how we prompt LLM w/ \texttt{Inter-Agent} to answer user queries with our proposed \textbf{\textsc{AIR}}.

\subsection{\textsc{Clarification} Action}
\begin{figure*}[t]
    \centering
    \includegraphics[width=0.9\textwidth]{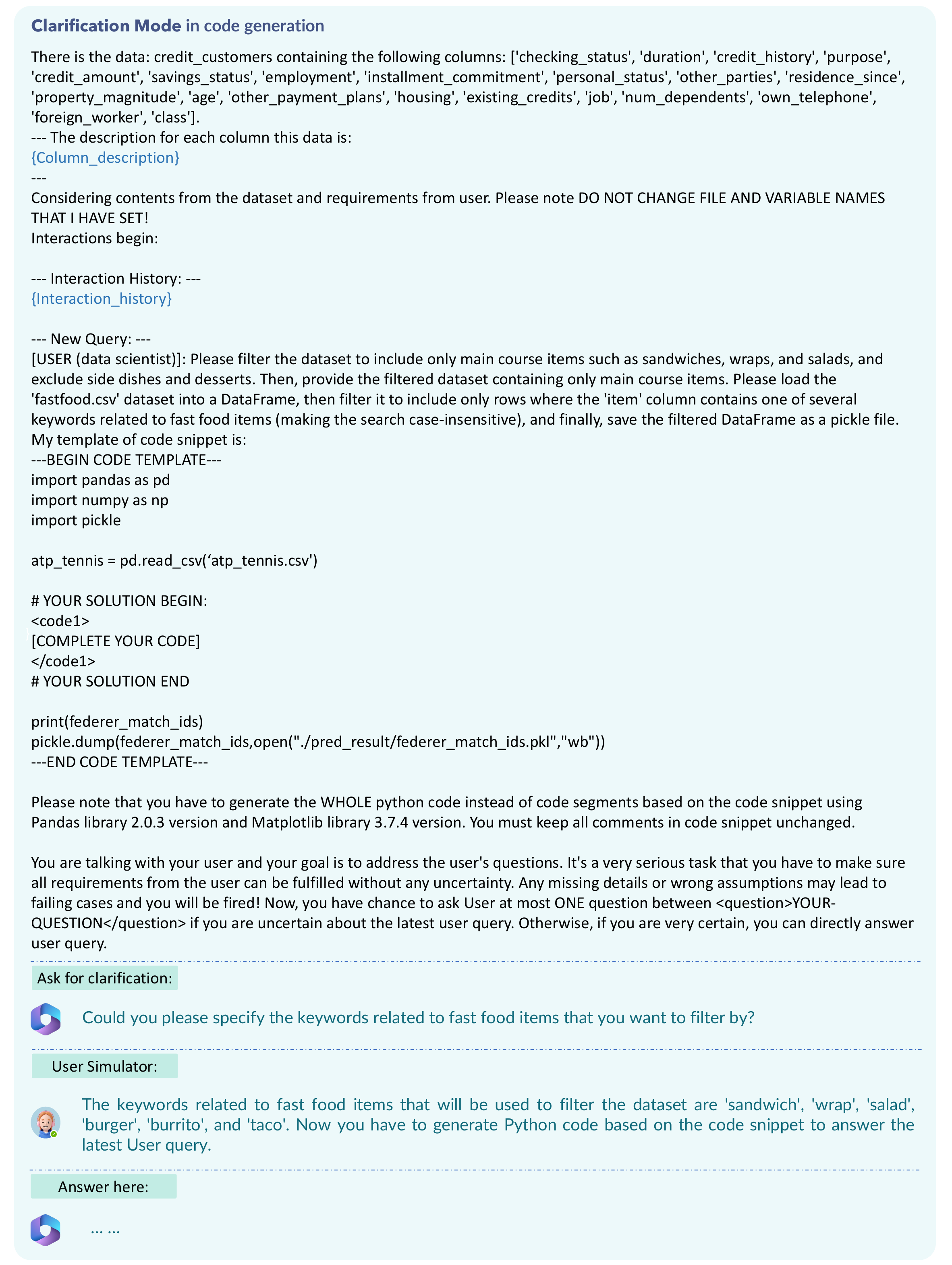}
    \caption{The prompt of LLM in \texttt{Model-Base} version in \textsc{clarification} mode.}
    \label{fig:prompt_clari}
\end{figure*}

The Figure \ref{fig:prompt_clari} describes how we prompt LLM in \texttt{Model-Base} version to ask for clarification.

\end{document}